\documentclass[lettersize,journal]{IEEEtran}
\usepackage{amsmath,amsfonts}
\usepackage{algorithmic}
\usepackage{algorithm}
\usepackage{array}
\usepackage[caption=false,font=normalsize,labelfont=sf,textfont=sf]{subfig}
\usepackage{textcomp}
\usepackage{stfloats}
\usepackage{url}
\usepackage{verbatim}
\usepackage{graphicx}
\usepackage{cite}

\usepackage{array}
\usepackage{multirow}
\usepackage{longtable}
\usepackage{rotating}
\usepackage{color}
\usepackage{arydshln}
\allowdisplaybreaks[3]
\newtheorem{theorem}{\bf Theorem}
\newtheorem{corollary}{\bf Corollary}
\newtheorem{lemma}{\bf Lemma}

\newtheorem{strategy}{\bf Strategy}
\newtheorem{assumption}{\bf Assumption}

\begin{document}

\title{CC-FedAvg: Computationally Customized Federated Averaging}

\author{Hao Zhang, Tingting Wu, Siyao Cheng, Jie Liu \IEEEmembership{Fellow, IEEE}
	\thanks{This work is partly supported by the National Key R\&D Program of China under Grant No. 2021ZD0110900, the National Natural Science Foundation of China under Grant No. 62106061, 61972114, the Fundamental Research Funds for the Central Universities under Grant No. FRFCU5710010521, the Research and Application of Key Technologies for Intelligent Farming Decision Platform, An Open Competition Project of Heilongjiang Province(China) under Grant No. 2021ZXJ05A03, the Key Research and Development Program of Heilongjiang Province under Grant No. 2022ZX01A22, the National Natural Science Foundation of Heilongjiang Province under Grant No. YQ2019F007.}
	\thanks{Corresponding author: Tingting Wu.}
	\thanks{Hao Zhang, Tingting Wu and Siyao Cheng are with the Faculty of Computing, Harbin Institute of Technology, Harbin 150000, China (e-mail: zhh1000@hit.edu.cn; ttw@ir.hit.edu.cn; csy@hit.edu.cn).}
	\thanks{Jie Liu is with the Institute for Artificial Intelligence, Harbin Institute of Technology (Shenzhen), Shenzhen 518055,  China (e-mail: jieliu@hit.edu.cn).}
}

\markboth{Journal of \LaTeX\ Class Files,~Vol.~X, No.~X, April~2023}%
{Shell \MakeLowercase{\textit{et al.}}: A Sample Article Using IEEEtran.cls for IEEE Journals}


\maketitle

\begin{abstract}
	Federated learning (FL) is an emerging paradigm to train model with distributed data from numerous Internet of Things (IoT) devices. It inherently assumes a uniform capacity among participants.  However, due to different conditions such as differing energy budgets or executing parallel unrelated tasks, participants have diverse computational resources in practice.  Participants with insufficient computation budgets must plan for the use of restricted computational resources appropriately, otherwise they would be unable to complete the entire training procedure, resulting in model performance decline. To address this issue, we propose a strategy for estimating local models without computationally intensive iterations. Based on it, we propose Computationally Customized Federated Averaging (CC-FedAvg), which allows participants to determine whether to perform traditional local training or model estimation in each round based on their current computational budgets. Both theoretical analysis and exhaustive experiments indicate that CC-FedAvg has the same convergence rate and comparable performance as FedAvg without resource constraints.  Furthermore, CC-FedAvg can be viewed as a computation-efficient version of FedAvg that retains model performance while considerably lowering computation overhead.
	
\end{abstract}

\section{Introduction}

With the rapid development of the Internet of Things (IoT), the number of IoT devices (e.g., smart phones, cameras and sensing devices) has increased dramatically. These devices are playing an increasingly important role in people's lives. They continually create large volumes of data, which encourages the use of artificial intelligence technology to improve a number of existing applications, including smart healthcare systems, automated driving, etc. However, due to the privacy concerns and communication limitations, keeping data locally is becoming increasingly appealing. Federated learning (FL)~\cite{mcmahan2017communication}, which allows multiple devices to collectively train a model without sharing local data, is emerging to make use of this dispersed data across a range of devices. 

Normally, a stable global model of FL is achieved after multiple rounds of local training and global aggregation. 
At each round, all or a portion of participants (usually selected by the server) calculate local models based on their local data, which is commonly iterated numerous times using optimization methods such as stochastic gradient descent (SGD) or its variations. 
This local training requires intensive on-device computing for long periods. However, devices in the wild greatly vary in computational resource budgets, resulting in some devices' computing resources being insufficient to complete training. For instance, devices with low energy budgets cannot provide enough computation capability to complete the entire training. 
Or other unrelated tasks that consume an arbitrary percentage of the resource, prevent the devices from providing sufficient computational resources for training. 
This ubiquitous presence of \emph{computation heterogeneity} is not considered by conventional FL methods, represented FedAvg. In fact, in both cross-silo and cross-device settings~\cite{kairouz2021advances}, all the clients have the same computational cost for conventional FL methods both in theoretical analysis and practical applications~\cite{reddi2020adaptive, yang2021achieving}, as the expected number of participation rounds remains constant.
Actually, devices with insufficient computational capacity cause straggler issues~\cite{mcmahan2017communication, yan2020distributed} 
(e.g., the devices are busy during training) or even  drop-out issues~\cite{lai2022fedscale, wang2022friends} (e.g., the battery is consumed during training), which slows down training and greatly harms performance~\cite{gu2021fast}. 


\begin{figure*}[t]
	\centering
	\includegraphics[width = 1.7\columnwidth]{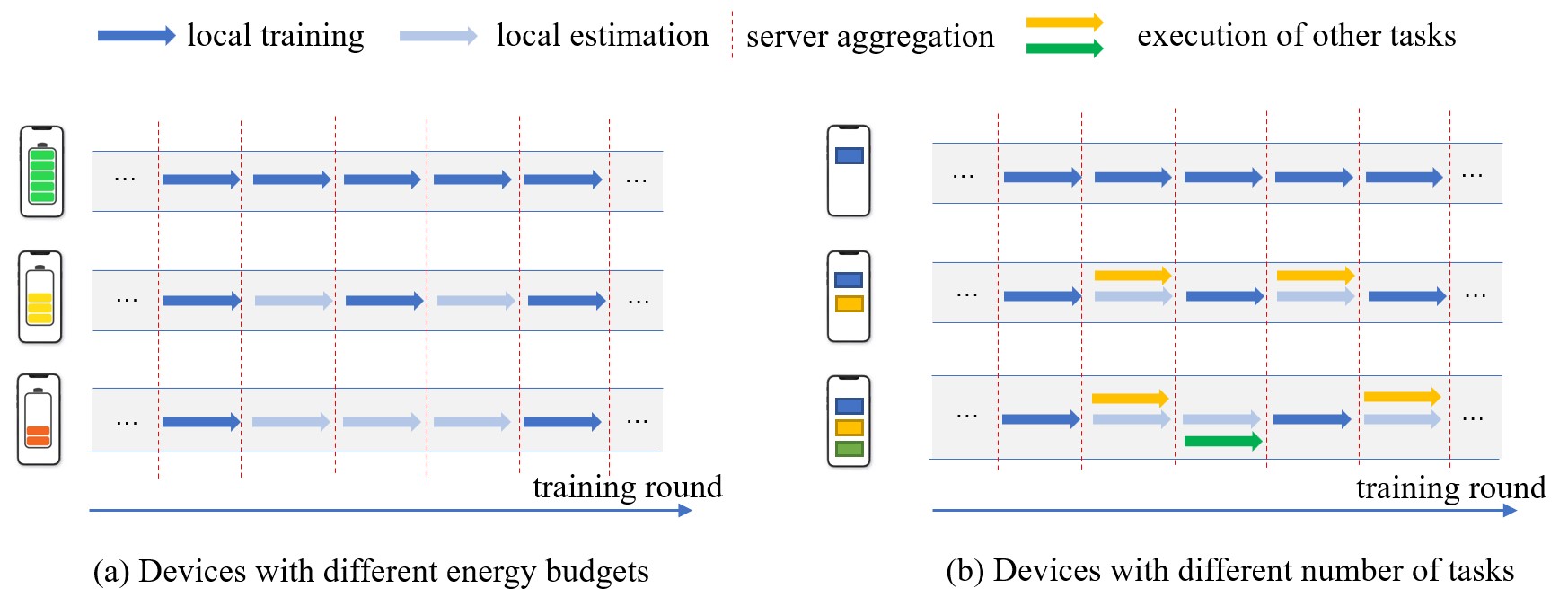} 
	\caption{Illustration of CC-FedAvg solving computation heterogeneity in FL training. ``Local training'' refers to the traditional iterative method such as SGD to obtain local models, which is computationally expensive. ``Local estimation'' refers to estimating the local model from historical information with negligible computational overhead.  (a) Devices schedule to train or estimate local models in advance based on their energy budgets. Devices with lower energy reserves perform local estimation with higher probability. (b) Devices decide whether to execute local training or local estimation based on the tasks in time. Devices with higher load perform local estimation with higher probability.}
	\label{fig:scenario} 
	\vskip -0.2in
\end{figure*}

To address this issue, devices would plan for the use of computational resource to guarantee throughout the training process. One approach is to reduce the number of local iterations for the clients with insufficient computation resources following~\cite{wang2020tackling}, but it only performs well in limited FL scenarios (see section~\ref{sec:nova} in detail). 
Another approach is to 
reduce the number of participating training rounds directly. 
In this way, the clients with insufficient computational resources skip certain training rounds proactively even when they are required to participate (e.g., selected by the server). 
As a result, the server would not receive the expected local models from these devices in the corresponding rounds. 
There are two intuitive strategies to deal with these devices. The first is to ignore them, aggregating global model only from the received local models (Strategy~\ref{strategy1}). This is how the vanilla FedAvg does~\cite{mcmahan2017communication}. However, it introduces bias into training data~\cite{yan2020distributed}, because the devices with sufficient computational resources participate more frequently, resulting in over-representation of their local data, and vice versa. It has been proved that biased selections lead to sub-optimal solutions, which have non-vanishing error gaps to the optimal one~\cite{cho2022towards, wang2022delta}. Moreover, if numerous devices skip the training, there will be insufficient participants, making the model's performance unstable~\cite{fraboni2021clustered}. The second is to utilize the latest local models of these devices (Strategy~\ref{strategy2}), so that all devices participate in training with the same number of rounds, forming an unbiased selection. However, this strategy uses stale local models, which would decrease the performance of the aggregated global model. 
As a result, the devices lacking of computational resources are required to \emph{provide new local models in corresponding rounds while avoiding substantial computational cost}.


To meet the requirement, we attempt to ``guess'' the local model with negligible computational overhead by clients themselves or even by the server, and we expect that it is as close as possible to the model obtained by training locally.
Actually, Strategy 2 can be regarded as an estimation method that estimates the local model in current round using the latest local model, although it differs markedly from the trained one. To get a better estimation, in this paper we propose a strategy for estimating a newly local model by historical information without computationally intensive iterations (Strategy~\ref{strategy3}).
We observe that the estimated model by Strategy~\ref{strategy3} is much closer to the trained one than that estimated by Strategy~\ref{strategy2}, especially at the early stage of training. Based on this strategy, we propose Computationally Customized Federated Averaging (CC-FedAvg), which can train FL models across clients with heterogeneous computational resources. CC-FedAvg allows the selected clients to determine whether to perform local training based on their computational resources at each training round (illustrated in Fig.~\ref{fig:scenario}). 
The main contributions of this paper can be summarized as follows:
\begin{itemize}
	\item We study the problem of computation heterogeneity in FL, and present a strategy for estimating the local model to be close to the one derived through local training. This estimation method results in negligible computational overhead. 
	\item Inspired by the preceding strategy, we propose CC-FedAvg, which allows clients with diverse computational resources to train FL model collaboratively. We prove that CC-FedAvg has the same convergence speed as vanilla FedAvg without limitation of computational resources (FedAvg (full) for short). 
	Our evaluation demonstrate that the performance of CC-FedAvg is almost the same as FedAvg (full), and much better than other two FL baselines and FedAvg with computation heterogeneity constraint.  
	\item Meanwhile, CC-FedAvg can be thought of a computation-efficient extension of FedAvg (full) that can save computational resources while retaining performance. It can replace FedAvg in general scenarios more than limiting to tackle computation heterogeneity. Experiments show that CC-FedAvg retains model performance even when the cost is reduced to $1/4$ for each participant.
\end{itemize}


\section{Related Work}
In practice the great variability of devices makes it challenging to deploy FL. Various kinds of heterogeneity need to be addressed. 

\textbf{Data heterogeneity.} 
Data heterogeneity is the most attractive topic, which starts with the birth of FL~\cite{mcmahan2017communication}. It is assumed that data are distributed across participants in a non-IID (Independent Identically Distributed) way, which results in large gaps among local models, aggregating a global model with poor performance~\cite{zhao2018federated}. Pioneer works like~\cite{karimireddy2020scaffold, liang2019variance, karimireddy2021breaking} correct the client drift actively to reduce the gaps among local models. Others like~\cite{li2018federated, yao2019federated, li2021model, zhang2022fedcos} introduce regularization to reduce the difference between local models.

\textbf{Device heterogeneity.}
In contrast to the significant work spent to data heterogeneity, tackling device heterogeneity has received less attention.
\cite{diao2021heterofl, horvath2021fjord, hong2022efficient, mei2022resourceadaptive}  address on system heterogeneity (mainly on run-time memory heterogeneity where participants are with different on-chip memory budgets). These methods enables devices to train size-adjustable local models based on their memory budgets, eliminating the constraint of standard FL methods that the global model complexity is limited for the most indigent device. Smaller models consume fewer computational resources to train, but at the expense of performance. 

At present computation heterogeneity is partly considered by some works. \cite{nishio2019client} handles the problem by selecting participants based on their resource information. However, a biased global model may be obtained because clients with more resources participate more. 
\cite{lin2020ensemble, zhou2020distilled} utilize model distillation for FL, where the architecture of local models can be decided based on participants' resources, but these methods usually rely on public datasets for knowledge transmission, which may not be realistic. 
Energy consumption is strongly correlated with computation overhead. \cite{luo2021cost, wang2019adaptive} minimize the total energy cost in FL training, whilst ignoring device heterogeneity. 
\cite{arouj2022towards} proposes an energy-aware FL selection method based on the uploaded participants' profiles to balance the energy consumption across these battery-constrained devices. \cite{kim2021autofl} exploits reinforcement learning to choose participants in each training round. However, these methods are still biased selection, resulting in a model that differs from the deterministic aggregation of all clients~\cite{fraboni2021clustered}.

Meanwhile, FedNova~\cite{wang2020tackling} can be regarded as a strategy of model estimation, using the model with inadequate training. It can be used to self-adapt the computational resource consumption of clients, but from our experiments, the way of reducing local iteration number in each round only works well in limited scenarios. Furthermore, in FedNova the server cannot help the participants to estimation as ours (Alg.~\ref{al:ccfl2} and Alg.~\ref{al:ccfl3} in Appendix), which lacks flexibility.   

\textbf{Training efficiency.}
Numerous studies try to speed up the training of FedAvg. 
SCAFFOLD~\cite{karimireddy2020scaffold} achieves comparable performance to FedAvg by altering the local models. CMFL~\cite{luping2019cmfl} analyzes local models before uploading to avoid upload irrelevant update. FedCos~\cite{zhang2022fedcos} introduces a cosine regularization term to reduce the directional inconsistency of local models. FedDyn~\cite{acar2021federated} uses linear and quadratic penalty terms to make local minima and global stationary point consistent. 
However, all of these methods incur more storage and calculation cost than FedAvg. Instead, our approach improves training effectiveness with almost any additional expense on FedAvg.


\section{Computationally Customized Federated Averaging}

\subsection{FL and Computation Heterogeneity}

We focus on federated training, in which $N$ clients are collaborated to yield a global model under the coordination of a parameter server. Formally, the goal is to solve the optimization problem as 
\small	
\begin{equation}
	\min_{x\in \mathbb{R}^d} f(x) = \frac{1}{N}\sum_{i=1}^N f_i(x),
\end{equation}
\normalsize	
where $f_i(x):=\mathbb{E}_{\xi_i \sim \mathcal{D}_i}[\ell_i(x,\xi_i)]$ is the local loss function of the $i$-th client with local dataset $\mathcal{D}_i$, and $x$ is model parameters. 
For normal federated learning methods such as FedAvg, the training process is divided into $T$ rounds. In the $t$-th round ($t<T$), the server randomly selects a part of clients (or all of them) denoted as $\mathcal{S}_t$  and broadcasts the global model $x_t$, which is aggregated in the latest round (for $t=0$, $x_0$ is randomly initialized by the server). For any client  in $\mathcal{S}_t$ (denote as client $i$), it initializes its local model $x_{t,0}^i = x_t$, and then performs $K$ Stochastic Gradient Descent (SGD) steps iteratively on local dataset with the learning rate $\eta$ as
\begin{equation}\label{eq:iteration}
\begin{small}	
	x^{i}_{t,k+1} = x^{i}_{t,k} - \eta g^i_{t,k},  k=0,1,\cdots, K-1,
\end{small}
\end{equation}
where $g^i_{t,k}$ is an unbiased estimator in each step, i.e., $\mathbb{E}[g^i_{t,k}] = \nabla f_i(x^{i}_{t,k})$.
After $K$ steps of iteration, the server aggregates updated local models by
\begin{equation}\label{eq:aggregation}
	\begin{small}
	x_{t+1} = \frac{1}{|\mathcal{S}_t|} \sum_{i\in \mathcal{S}_t}  x^{i}_{t,K}.
\end{small}
\end{equation}
In the traditional FL process, all clients consume the same amount of computational resources in expectation. In actuality, however, they may not have the same computational resource budgets. This is referred to as \emph{computation heterogeneity}. For example, due to lack of energy reserves, some clients are unable to execute the same amount of computation as others. Even in the absence of energy constraints, the imbalance of computing resources still exists. For instance, while performing federated training, the client may process multiple other tasks concurrently. Because of the resource preemption, the computational resources for some participants in federated training are insufficient. 
As a result, these clients with insufficient computational resources are unable to complete $K$ iterations before a predetermined time in one round. They would be regarded as ``stragglers'', even drop out the training~\cite{wang2022friends}.


\begin{figure*}[ht]
	\centering
	\subfloat[Euclidean distance (the smaller the better)]{
		\includegraphics[width=0.65\columnwidth]{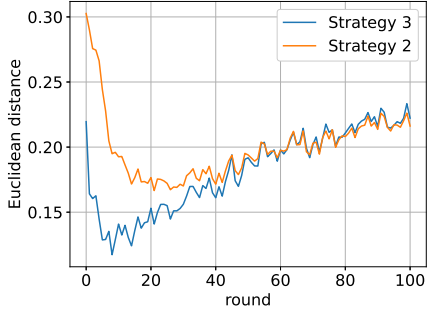} 
		\label{fig:euclidean} 
	}
	\quad \quad
	\subfloat[Cosine distance (the larger the better)]{
		\includegraphics[width=0.67\columnwidth]{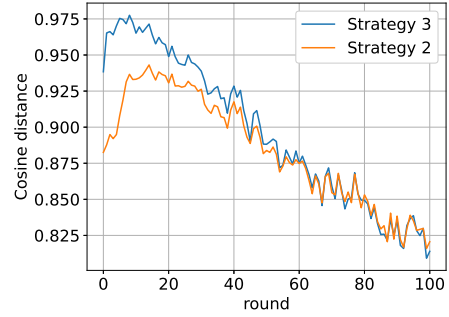} 
		\label{fig:cosine} 
		
	}
	\caption{\textcolor{black}{We investigate the deviation between the estimated models by different strategies and the true models derived through local training for one random chosen client (indexed by $i$) on CIFAR-10. It indicates the accuracy of the estimation. In (a) we measure the Euclidean distances between the estimated models ($x^i_{t-1,K}$ by Strategy~\ref{strategy2} and $(x^i_{t,0} + \Delta_{t-1}^i)$ by Strategy~\ref{strategy3}) and the true model ($x^i_{t,K}$) in each round. For Strategy~\ref{strategy2}, the distance is $\| x^i_{t,K} - x^i_{t-1,K}\|^2$, where $t$ is the round index. For Strategy~\ref{strategy3}, the distance is $\| x^i_{t,K} - (x^i_{t,0} + \Delta_{t-1}^i)\|^2$. In (b) we measure the deviation of movement from the initial model in each round by cosine distance. For Strategy~\ref{strategy2}, it is $\left<\Delta^i_{t},  x^i_{t-1,K}-x^i_{t,0}\right>$. For Strategy~\ref{strategy3}, it is $\left<\Delta^i_{t},  \Delta^i_{t-1}\right>$. Both measurements indicate that the model estimated by Strategy~\ref{strategy3} is a competitive estimation compared with the model by Strategy~\ref{strategy2}, especially in the early stage of training.}}
	\label{fig:distance} 
	\vskip -0.2in
\end{figure*}

\subsection{Motivation and Baselines}

To address this issue, we should balance computing overhead among clients with heterogeneous computational resources, i.e., we should reduce the computation load of clients with insufficient computational resources to avoid bottlenecks.
Formally, for client $i$, 
the ratio of remaining computational resources (compared with the required computational resources in traditional FL) is denoted as $p_i \in (0,1]$. 
The scarcer the computational resource of client is, the smaller $p_i$ is.
Particularly, if the client has sufficient resources, $p_i=1$.  
Since the computational resources of clients are mainly consumed in the iterations of local gradient updates, the intuitive way to lower computational overhead is to reduce the number of participating rounds. 
Specifically, for round $t$, if client $i$ is in $\mathcal{S}_t$, it skips local iterations in this round with probability $1-p_i$. 
In the entire training process, if all clients are selected by the server in each round, client $i$ calculates gradients by $p_i\times K\times T$ times (while it needs $K\times T$ times in traditional FL). 
Hence the computation cost is reduced by $1-p_i$ totally.

The server needs to deal with the clients skipping local training in the aggregation phase.  
Accordingly, there are two intuitive ways to deal with these clients. The first is to ignore them directly, forming the strategy as follows. 

\begin{strategy}\label{strategy1}
	If client $i$ will skip round $t$, it sends back ``skip" signal to the server (or sends nothing at all). The server removes it from $\mathcal{S}_t$ in aggregation phase when receiving the signal (or waiting until timeout).
\end{strategy}

This strategy is how vanilla FedAvg 
does. Strategy~\ref{strategy1} ensures that the computing consumption meets the requirement of  computation heterogeneity, but it results in a biased client sampling scheme, leading to the aggregated model to be different from the deterministic aggregation of all clients. It makes the model sub-optimal~\cite{cho2022towards}. 
To conquer this problem, the second strategy requires all clients in $\mathcal{S}_t$ to take part in aggregation phase, regardless of whether local training is skipped.
 
\begin{strategy}\label{strategy2}
If client $i$ will skip round $t$, it returns the local model of the last round $x^{i}_{t-1,K}$ to the server instead of calculating  $x^{i}_{t,K}$.
\end{strategy}

This strategy is also presented by previous literature such as~\cite{yan2020distributed}. However, stale local models are used in Strategy~\ref{strategy2}, which reduces the performance of aggregated model. 

To overcome the drawbacks of both strategies, the clients that skip current round should provide newly local models but without computationally intensive local training. 
Although it appears contradictory, it motivates us 
to \emph{``guess" a local model that is as close as possible to the model obtained by gradient iterations, but with negligible computational overhead}. Indeed, Strategy~\ref{strategy2} can be viewed as a model estimation, but the estimation is insufficiently precise,  resulting in poor performance.

For client $i$, we observe the movement of the local model in round $t-1$ and get the moving vector $\Delta^i_{t-1} = x^i_{t-1,K} - x^i_{t-1,0}$. In the subsequent round $t$, we assume that the local model would do the same movement as before. Hence we have following strategy to estimate the local model:

\begin{strategy}\label{strategy3}
If client $i$ will skip round $t$, it returns the estimated local model $x^i_{t,0} + \Delta^i_{t-1}$ as $x^{i}_{t,K}$.
\end{strategy}


Figure~\ref{fig:distance} compares the model estimations of Strategy~\ref{strategy3} and Strategy~\ref{strategy2} by both Euclidean distance and cosine distance. In Fig.~\ref{fig:euclidean} we illustrate the Euclidean distances between observed local models and the true model inferred by Eq.~(\ref{eq:iteration}). For Strategy~\ref{strategy2}, the estimated model is the one obtained in the previous round. In Fig.~\ref{fig:cosine} we measure the cosine distances between the moving vector from $x^{i}_{t,0}$ to estimated models and to the true model respectively. 
\textcolor{black}{Both measurements show that the model observed by Strategy~\ref{strategy3} is close to the real one that is obtained by local training, and is a competitive estimation compared with the model by Strategy~\ref{strategy2}, especially in the early stage of training. Therefore, compared with Strategy~\ref{strategy2}, Strategy~\ref{strategy3} is a better one to estimate the local model.}

\subsection{Method}\label{sec:method}

\begin{algorithm}[htb]
	\caption{\textbf{CC-FedAvg}: Computationally Customized Federated Averaging 
	} 
	\label{al:ccfl}
	\begin{algorithmic}[1]
		\STATE Initialize $x_0$.
		\FOR {$t = 0$ to $T-1$}
		\STATE Server selects $\mathcal{S}_t$ randomly and sends $x_{t}$ to clients in $\mathcal{S}_t$.
		\STATE (At client:)\\
		\FOR {each client $i \in \mathcal{S}_t$ in parallel}
		\IF{not skip this round} 
		\STATE /* with probability $p_i$ (determined in advance or in real time based on local computational resource) */
		\STATE initial local model: $x_{t,0}^i = x_{t}$.\\ 
		\FOR {$k=0$ to $K-1$}
		\STATE Update local model by an unbiased estimate of gradient: $x^i_{t,k+1} = x^i_{t,k} - \eta g_{t,k}^i$.\\
		\ENDFOR
		\STATE Get $\Delta^i_t = x^i_{t,K} - x^i_{t,0}$.\label{line:ltend}\\
		\ELSE 
		\STATE /* estimate local model (with probability $1-p_i$)*/
		\STATE Get $\Delta^i_t = \Delta^i_{t-1}$.\label{line:skip}\\
		\ENDIF
		\STATE Send $\Delta^i_t$ back to the server.\\
		\ENDFOR
		\STATE (At server:)\\
		\STATE $\Delta_t = \frac{1}{|\mathcal{S}_t|}\sum_{i \in \mathcal{S}_t} \Delta^i_t$.\\
		\STATE Update global model $x_{t+1} = x_t + \Delta_t$.\\
		\ENDFOR
		\vskip -0.2in
	\end{algorithmic}
\end{algorithm}

Inspired by the observation, we propose Computational Customized Federated Averaging (CC-FedAvg) illustrated in Algorithm~\ref{al:ccfl}. The process is similar as vanilla FedAvg. The different is that the client decides to either perform local training as vanilla FedAvg 
(line~8 to line~12) or estimate the local update 
(line~15) according to its computational resources in each round, if it is selected by the server. From the whole training process, each client gets local model by performing local training with probability $p_i$ and by estimation with probability $1-p_i$, where $p_i$ is determined by the adequacy of its local computing resources. 
Actually, for client $i$ with $p_i < 1$, more than one successive rounds may be skipped, leading to $\Delta_t^i = \Delta_{t-1}^i = \Delta_{t-2}^i = \cdots$. 

Obviously, if the client skips one round by estimating the movement of its local model, there is almost no computational overhead compared with computation-consuming local training. 
Therefore, smaller $p_i$ means lower computational overhead during the training. Each client can decide $p_i$ by itself in advance or in real time to adapt local resources. 
Accordingly, vanilla FedAvg can be regarded as a special case of CC-FedAvg where all clients with $p_i=1$, i.e., all clients have sufficient resources. 
Meanwhile, model estimation 
(line~12) 
also can be executed on the server, which further reduces the requirement to local storage cost and communication cost. In Appendix~\ref{sec:variantsCC-FedAvg} we propose other two variants of CC-FedAvg. Because all variants are based on the same principle, in the following we analyze CC-FedAvg based on Algorithm~\ref{al:ccfl} by default.

\textcolor{black}{It is worth noting that \cite{gu2021fast} proposes a method named MIFA similar as ours, especially as Alg.~\ref{al:ccfl2} where the server calculates the local estimation of devices. However, there are several differences from ours. First, they have different goals. MIFA aims to deal with straggling devices, where the scenario still is computation homogeneity. Hence the motivation, theoretical analysis and experiments are all different with ours. MIFA does not consider the influence of the ratio of insufficient computational resources as ours, which is very critical in heterogeneous scenarios. Second, MIFA complements the local model of inactive clients. It is equivalent to requiring all participants to join the aggregation, changing the``cross-device'' scenario into the ``cross-silo'' scenario, which may introduces too much stale information. In our method, we still follow the scenarios of original FL, only considering the selected participants in each round. Third, MIFA also considers the situation of providing estimation from participants themselves, but it aims to solving the influence of straggling by keeping the stale information of clients passively. On the contrary, CC-FedAvg aims to reducing computing overhead, where the participants upload model estimations proactively according to its conditions. At last, we aim to estimate local models without expensive operations.  Line~\ref{line:skip} is a representative estimation based on  Strategy~\ref{strategy3}. It also can be replaced by other estimations totally different from MIFA. For example, Fig.~\ref{fig:distance} shows that the local model by Strategy~\ref{strategy2} is as close to the true one as the model by Strategy~\ref{strategy3} after several epochs. It inspires a combination of Strategy~\ref{strategy2} and Strategy~\ref{strategy3}. Correspondingly, the estimation of line~\ref{line:skip} can be replaced by 
\begin{equation}\label{conbine}
	\Delta_t^i = \left\{
	\begin{array}{lcll}
		\Delta_{t-1}^i,  & & t < \tau &  \text{(Strategy~\ref{strategy3})} \\
		x_{t-1,K}^i-x_{t,0}^i,  & &   t \geq \tau &   \text{(Strategy~\ref{strategy2})}
	\end{array}
	\right.
\end{equation}
where $\tau$ is a threshold of epoch number that determines whether to use Strategy~\ref{strategy2} or Strategy~\ref{strategy3} to estimate the local model. $x_{t-1,K}^i$ is the latest local models of client $i$. In section \ref{sec:estimation} we compare the performance of the replacement.}

\section{Theoretical Analysis}

\subsection{Convergence Analysis}
In order to analyze the convergence of CC-FedAvg, we first state the assumptions as follows.

\begin{assumption}[L-Lipschitz Continuous Gradient]\label{assum:smooth}
\textit{Function $\displaystyle f_i(x)$ is $L$-smooth, i.e., $\forall x,y \in \mathbb{R}^d$,
\begin{small}
\begin{equation}\label{eq:smooth}
	\|\nabla f_i(y) - \nabla f_i   (x)\|  \leq  L\| y-x\|
\end{equation}
\end{small}} 
\end{assumption}

\begin{assumption}[Unbiased Local Gradient Estimator]\label{assum:unbias}
	\textit{For the $i$-th client, the local gradient estimator $g^i_{t,k}=\nabla f_i(x_{t,k}^i, \xi^i_{t,k})$ is unbiased and the variance is bounded, where $\xi^i_{t,k}$ is a random mini-batch of local data in the $k$-th step of the $t$-th round at the $i$-th client, i.e., 
	\begin{small}
	\begin{equation}\label{eq:meanequal}
		\mathbb{E}[g^i_{t,k}] = \nabla f_i   (x_{t,k}^i) 
	\end{equation}
	\begin{equation}\label{eq:localvar}
	\|g^i_{t,k} - \nabla f_i   (x_{t,k}^i)\|^2  \leq  \sigma^2_L
	\end{equation}
	\end{small}}
\end{assumption}

\begin{assumption}[Bounded Global Variance]\label{assum:gvar}
	\textit{The variance of local gradients with respect to the global gradient is bounded, i.e.,
	\begin{small}
	\begin{equation}\label{eq:globalvar}
		\displaystyle  \mathbb{E} \| \nabla f_i(x) - \nabla f(x) \|^2 \leq \sigma^2_G.
	\end{equation}
	\end{small}}
\end{assumption}

Assumption~\ref{assum:smooth} and~\ref{assum:unbias} are standard in general non-convex optimization~\cite{Jorge2016Optimization,wang2021field,haddadpour2019convergence,yin2018gradient,Zhang2023}. Assumption~\ref{assum:gvar} is also a common assumption in federated optimization~\cite{yang2021achieving,reddi2021adaptive,ijcai2019-637}, where $\sigma_G$ is used to quantify the degree of data heterogeneity among clients. In particular, $\sigma_G = 0$ corresponds to IID data setting. 

Based on the above assumptions, we present the theoretical results for the non-convex problem. For simplicity, we consider CC-FedAvg with full client participation, i.e. $|\mathcal{S}_t|=N$.

\begin{theorem}\label{th:nonconvex}
\textit{For $N$ participants, suppose there are $r\cdot N$ clients $(r \in [0,1])$ with insufficient computational resources, each of which performs local iterations at least once every $W$ rounds ($W \ll T$). If the local learning rate satisfies $\eta  \leq \min{(\frac{\sqrt{1+24r^2W^2L}-1}{12r^2KW^2L}, \frac{1}{\sqrt{60(6r^2+1)}KL})}$, we have the following convergence result for CC-FedAvg:
\begin{equation}\label{eq:nonconvex}
	\begin{small}
	\begin{aligned}
		\min_{t\in[T]} \|\nabla f(x_t)\|^2 \leq \frac{f(x_0) - f(x^*)}{c\eta K T} + \frac{D}{c\eta K}, 
	\end{aligned}
\end{small}
\end{equation} 
where $c$ is a constant satisfying $0<c<\frac{1}{2}$ and
\begin{small}
\begin{equation}
	\begin{aligned}
		D = 5(6r^2 + 1) (\sigma_L^2+ 6K\sigma_G^2) L^2 K^2\eta^3 + \frac{K\eta^2L}{2N}\sigma_L^2.
	\end{aligned}
\end{equation}
\end{small}}
\end{theorem}

\begin{IEEEproof}
For the $t$-th round,
	\begin{equation}\label{eq:oneround}
		\begin{aligned}
			\mathbb{E}[f(x_{t+1})] &\leq f(x_t) + \langle\nabla f(x_t), \mathbb{E}[x_{t+1} - x_{t}]\rangle \\&+ \frac{L}{2}\mathbb{E}\|x_{t+1} - x_{t}\|^2 \\
			&\overset{(a)}{=}f(x_t) + \langle\nabla f(x_t), \mathbb{E}[\Delta_t+\eta K\nabla f(x_t) \\&-\eta K\nabla f(x_t)]\rangle + \frac{L}{2}\mathbb{E}\|\Delta_t\|^2 \\
			&= f(x_t) -\eta K\|\nabla f(x_t)\|^2 \\&+ \underbrace{\langle\nabla f(x_t), \mathbb{E}[\Delta_t]+\eta K\nabla f(x_t)\rangle}_{A_1} + \frac{L}{2}\mathbb{E}\|\Delta_t\|^2.
		\end{aligned}
	\end{equation}
	The term $A_1$ can be bounded as follows
	\begin{equation}
		\begin{aligned}
			A_1 =&\langle\sqrt{\eta K}\nabla f(x_t), \frac{1}{\sqrt{\eta K}}\mathbb{E}[\Delta_t]+\sqrt{\eta K}\nabla f(x_t)\rangle \\
			\overset{(a_1)}{=}&\frac{\eta K}{2}\|\nabla f(x_t)\|^2 + \frac{1}{2\eta K}\| \mathbb{E}[\Delta_t]+\eta K\nabla f(x_t)\|^2 \\&- \frac{1}{2\eta K}\| \mathbb{E}[\Delta_t]\|^2
		\end{aligned}
	\end{equation}
	where ($a_1$) follows $2\langle a, b\rangle = \|a\|^2+\|b\|^2-\|a-b\|^2$. 
	Thus 
	\begin{equation}
		\begin{aligned}
			\mathbb{E}[f(x_{t+1})] &\leq f(x_t) - \frac{\eta K}{2}\|\nabla f(x_t)\|^2 + \frac{1}{2\eta K}\| \mathbb{E}[\Delta_t] \\&+\eta K\nabla f(x_t)\|^2 \\
			&\quad\quad + \frac{L}{2}\mathbb{E}\|\Delta_t\|^2 - \frac{1}{2\eta K}\| \mathbb{E}[\Delta_t]\|^2 \\
			&\leq f(x_t) - \frac{\eta K}{2}\|\nabla f(x_t)\|^2 \\&+ \underbrace{\frac{1}{2\eta K}\| \mathbb{E}[\Delta_t]+\eta K\nabla f(x_t)\|^2}_{A_2} \\
			&\quad\quad + \frac{K\eta^2L}{2N}\sigma_L^2 - (\frac{1}{2\eta K} - \frac{L}{2}) \| \mathbb{E}[\Delta_t]\|^2
		\end{aligned}
	\end{equation}

	For $A_2$:
	\begin{equation}\label{eq:enlarge}
		\begin{aligned}
			&\frac{1}{2\eta K}\| \mathbb{E}[\Delta_t]+\eta K\nabla f(x_t)\|^2 \\
			=& \frac{1}{2\eta K}\left\|-\frac{\eta}{N}\sum_{i=1}^N\sum_{k=0}^{K-1} \nabla f_i(x_{t,k}^i) \right. \\& \left.+ \frac{\eta}{N}\sum_{j=1}^W\sum_{i \in \mathcal{M}_j}\sum_{k=0}^{K-1} \big(\nabla f_i(x_{t,k}^i) -\nabla f_i(x_{t-j,k}^i)\big) \right. \\
			&\left. \quad  + \frac{\eta}{N}\sum_{i=1}^N\sum_{k=0}^{K-1} \nabla f_i(x_{t})\right\|^2 \\
			\overset{(a_2)}{\leq}& \underbrace{\frac{\eta}{K} \left\| \frac{1}{N}\sum_{j=1}^W\sum_{i \in \mathcal{M}_j}\sum_{k=0}^{K-1} \big(\nabla f_i(x_{t,k}^i) -\nabla f_i(x_{t-j,k}^i)\big) \right\|^2}_B \\
			&\quad +\underbrace{\frac{\eta}{K} \left\| \frac{1}{N}\sum_{i=1}^N\sum_{k=0}^{K-1} \nabla f_i(x_{t,k}^i) - \frac{1}{N}\sum_{i=1}^N\sum_{k=0}^{K-1} \nabla f_i(x_{t})\right\|^2}_C 
		\end{aligned}
	\end{equation}
	where ($a_2$) follows $\|a+b\|^2 \leq 2\|a\|^2+2\|b\|^2$,

	The term $B$ is bounded as 
		\begin{align*}
			B=&\frac{\eta}{K} \left\| \frac{1}{N}\sum_{j=1}^W\sum_{i \in \mathcal{M}_j}\sum_{k=0}^{K-1} \big(\nabla f_i(x_{t,k}^i) -\nabla f_i(x_{t-j,k}^i)\big) \right\|^2 \\
			\overset{(b_1)}{\leq}& \frac{M\eta }{N^2} \sum_{j=1}^W\sum_{i \in \mathcal{M}_j}\sum_{k=0}^{K-1} \left\|\big(\nabla f_i(x_{t,k}^i) -\nabla f_i(x_{t-j,k}^i)\big) \right\|^2 \\
			\overset{(b_2)}{\leq}& \frac{M\eta L^2}{N^2} \sum_{j=1}^W\sum_{i \in \mathcal{M}_j}\sum_{k=0}^{K-1} \left\|x_{t,k}^i -x_{t-j,k}^i \right\|^2 \\
			\overset{(b_3)}{\leq}& \frac{3M\eta L^2}{N^2} \Big(
			\underbrace{\sum_{j=1}^W\sum_{i \in \mathcal{M}_j}\sum_{k=0}^{K-1} \left\|x_{t,k}^i -x_{t} \right\|^2}_{B_1} \\&+ 
			\underbrace{\sum_{j=1}^W\sum_{i \in \mathcal{M}_j}\sum_{k=0}^{K-1} \left\|x_{t} -x_{t-j} \right\|^2}_{B_2} \\
			&\quad\quad\quad\quad\quad\quad+ \underbrace{\sum_{j=1}^W\sum_{i \in \mathcal{M}_j}\sum_{k=0}^{K-1} \left\|x_{t-j,k}^i -x_{t-j} \right\|^2}_{B_3}\Big)
		\end{align*}
	where ($b_1$) and ($b_3$) follow $\|\sum_{i=1}^n a_i\|^2 \leq n\sum_{i=1}^n \|a_i\|^2$, ($b_2$) follows Assumption~\ref{assum:smooth}.
	
	For the three items $B_1$, $B_2$ and $B_3$, 
	\begin{equation}
		\begin{aligned}
			B_1 &= \sum_{j=1}^W\sum_{i \in \mathcal{M}_j}\sum_{k=0}^{K-1} \left\|x_{t,k}^i -x_{t} \right\|^2 
			\leq M(H_1+H_2 \|\nabla f(x_t)\|^2) \\
		\end{aligned}
	\end{equation}
	\begin{equation}
		\begin{aligned}
			B_2 &= \sum_{j=1}^W\sum_{i \in \mathcal{M}_j}\sum_{k=0}^{K-1} \left\|x_{t} -x_{t-j} \right\|^2 = K\sum_{j=1}^W m_j \left\|x_{t} -x_{t-j} \right\|^2\\
			&= K\sum_{j=1}^W m_j \| \sum_{s=0}^{j-1} (x_{t-s} - x_{t-s-1}) \|^2 \\
			&\overset{(b_4)}{\leq} K\sum_{j=1}^W m_j \cdot j\sum_{s=0}^{j-1}\| x_{t-s} - x_{t-s-1} \|^2 \\
			&\overset{(b_5)}{=} K\sum_{j=1}^W m_j \cdot j\sum_{s=0}^{j-1}\| \mathbb{E}[\Delta_{t-s-1}] \|^2 \\
			&\overset{(b_6)}{\leq} KWM \sum_{j=1}^W \|\mathbb{E}[\Delta_{t-j}]\|^2,
		\end{aligned}
	\end{equation}
	where ($b_4$) follows $\|\sum_{i=1}^n a_i\|^2 \leq n\sum_{i=1}^n \|a_i\|^2$, ($b_5$) holds since there are only the items in the form of $\nabla f_i(x_{t,k}^i)$ without the form $g_{t,k}^i)$,  we can use $\| \mathbb{E}[\Delta_{t-s-1}] \|^2$ instead of $\mathbb{E}\| \Delta_{t-s-1} \|^2$, ($b_6$) follows $j\leq W$.
	
	\begin{equation}
		\begin{aligned}
			B_3 &= \sum_{j=1}^W\sum_{i \in \mathcal{M}_j}\sum_{k=0}^{K-1} \left\|x_{t-j,k}^i -x_{t-j} \right\|^2 
			\\& \leq \sum_{j=1}^W m_j(H_1+H_2 \|\nabla f(x_{t-j})\|^2)
		\end{aligned}
	\end{equation}
	
	The term $C$ is bounded as 
	\begin{equation}
		\begin{aligned}
			C=&\frac{\eta}{K} \left\| \frac{1}{N}\sum_{i=1}^N\sum_{k=0}^{K-1} \nabla f_i(x_{t,k}^i) - \frac{1}{N}\sum_{i=1}^N\sum_{k=0}^{K-1} \nabla f_i(x_{t})\right\|^2 \\
			\overset{(c_1)}{\leq}& \frac{\eta}{N}\sum_{i=1}^N\sum_{k=0}^{K-1}\|\nabla f_i(x_{t,k}^i) - \nabla f_i(x_{t})\|^2 \\
			\overset{(c_2)}{\leq}& \frac{\eta L^2}{N} \sum_{i=1}^N\sum_{k=0}^{K-1} \| x_{t,k}^i - x_{t}\|^2 \\
			\overset{(c_3)}{\leq}& \eta L^2 (H_1+H_2\|\nabla f(x_t)\|^2)
		\end{aligned}
	\end{equation}
	where ($c_1$) follows $\|\sum_{i=1}^n a_i\|^2 \leq n\sum_{i=1}^n \|a_i\|^2$, ($c_2$) follows Assumption~\ref{assum:smooth}, and ($c_3$) follows Lemma~\ref{lemma:movementsum}.
	
	Therefore, by putting the pieces together into Eq.~(\ref{eq:oneround}), we obtain
	\begin{equation}
		\begin{aligned}
			&\mathbb{E}[f(x_{t+1})] \\
			\leq& f(x_t) - \frac{\eta K}{2}\|\nabla f(x_t)\|^2 + \frac{3M\eta L^2}{N^2}\Big(M(H_1+H_2\|\nabla f(x_t)\|^2) \\& + KWM \sum_{j=1}^W \|\mathbb{E}[\Delta_{t-j}]\|^2 + \sum_{j=1}^W m_j(H_1+H_2 \|\nabla f(x_{t-j})\|^2)\Big) \\& + \eta L^2 (H_1+H_2\|\nabla f(x_t)\|^2) + \frac{K\eta^2L}{2N}\sigma_L^2 \\& - (\frac{1}{2\eta K} - \frac{L}{2}) \| \mathbb{E}[\Delta_t]\|^2. 
		\end{aligned}
	\end{equation}
	We rewrite it as
	\begin{equation}
		\begin{aligned}
			&\Big(\frac{\eta K}{2}  
			- (\frac{3M^2\eta L^2}{N^2} + \eta L^2)H_2\Big)\|\nabla f(x_t)\|^2 \\& 
			- \frac{3M\eta L^2}{N^2}\sum_{j=1}^W m_j H_2 \|\nabla f(x_{t-j})\|^2 \\
			\leq&  f(x_t) - \mathbb{E}[f(x_{t+1})] - (\frac{1}{2\eta K} - \frac{L}{2}) \| \mathbb{E}[\Delta_t]\|^2 \\ 
			&+ \frac{3M^2K\eta W L^2}{N^2} \sum_{j=1}^W \|\mathbb{E}[\Delta_{t-j}]\|^2 \\ 
			&+ (\frac{6M^2\eta L^2}{N^2} + \eta L^2) H_1 + \frac{K\eta^2L}{2N}\sigma_L^2.
		\end{aligned}
	\end{equation}
	Here we \textcolor{black}{obtain} the upper bound of the $t$-th round. Then we rearrange the above inequalities from $t=0$ to $T-1$ and average them, and denote the virtual items of $t=-1, -2, \cdots$ (e.g., $\|\nabla f(x_{-1})\|^2$, $\| \mathbb{E}[\Delta_{-1}]\|^2$, etc.) are equal to $0$. Meanwhile, 
	suppose $\eta \leq \frac{N}{\sqrt{60(6M^2+N^2)}KL}$, there exists a constant $c$ satisfying $0 < c < \big(\frac{1}{2} - (\frac{6M^2 L^2}{N^2} +  L^2)\frac{H_2}{K}\big)$. For a sufficiently large $T$ and $K \ll T$, we have
	\begin{equation}
		\begin{aligned}
			&c\eta K\frac{1}{T}\sum_{t=0}^{T-1}  \|\nabla f(x_t)\|^2 \\	
			\overset{(d_1)}{\leq}&\frac{1}{T}\sum_{t=0}^{T-1} \big(\frac{\eta K}{2}  
			- (\frac{6M^2\eta L^2}{N^2} + \eta L^2)H_2\big) \|\nabla f(x_t)\|^2 \\
			{\leq}& \frac{1}{T}(f(x_0) - f(x_T))  
			+ (\frac{6M^2\eta L^2}{N^2} + \eta L^2) H_1 + \frac{K\eta^2L}{2N}\sigma_L^2  \\
			&- \frac{1}{T}\sum_{t=0}^{T-1}(\frac{1}{2\eta K} - \frac{L}{2} - \frac{3M^2K\eta W^2 L^2}{N^2}) \| \mathbb{E}[\Delta_t]\|^2 \\
			\overset{(d_2)}{\leq}& \frac{1}{T}(f(x_0) - f(x_T)) + (\frac{6M^2\eta L^2}{N^2} + \eta L^2) H_1 + \frac{K\eta^2L}{2N}\sigma_L^2, 
		\end{aligned}
	\end{equation}
	where ($d_1$) holds for $\eta \leq \frac{N}{\sqrt{60(6M^2+N^2)}KL}=\frac{1}{\sqrt{60(6r^2+1)}KL}$, ($d_2$) holds for $\eta \leq \frac{(\sqrt{N^2+24M^2W^2L}-N)N}{12M^2KW^2L}=\frac{\sqrt{1+24r^2W^2L}-1}{12r^2KW^2L}$ where the items containing $\| \mathbb{E}[\Delta_t]\|^2$ is eliminated.   

	Hence we obtain
	\begin{equation}
		\begin{aligned}
			min_{t\in[T]} \|\nabla f(x_t)\|^2 \leq& \frac{1}{T}\sum_{t=0}^{T-1}  \|\nabla f(x_t)\|^2 \\
			{\leq}& \frac{f(x_0) -  f(x^*)}{c\eta K T} + \frac{D}{c\eta K} 
		\end{aligned}
	\end{equation} 
	where 
	\begin{equation}
		\begin{aligned}
			D =& 5(\frac{6M^2 }{N^2} + 1) (\sigma_L^2+ 6K\sigma_G^2) L^2 K^2\eta^3 + \frac{K\eta^2L}{2N}\sigma_L^2 \\
			=& 5(6r^2 + 1) (\sigma_L^2+ 6K\sigma_G^2) L^2 K^2\eta^3 + \frac{K\eta^2L}{2N}\sigma_L^2.
		\end{aligned}
	\end{equation}
\end{IEEEproof}

For client $i$, $W$ is strongly related to $p_i$. If the client participates in FL in a round-robin manner, i.e., once per $W$ rounds (``round-robin'' schedule in Section~\ref{settings}), $p_i=\frac{1}{W}$. If the client participates at random, i.e., once randomly in $W$ rounds (``ad-hoc'' schedule in Section~\ref{settings}), $p_i=\mathbb{E}[\frac{1}{W}]$. 
As a result, we view $\frac{1}{W}$ as being equivalent to $p_i$ by expectation.

With Theorem~\ref{th:nonconvex}, we have the following convergence rate for CC-FedAvg with a proper choice learning rate:
\begin{corollary}\label{cor:converagerate}
	\textit{Let $\eta=\frac{\sqrt{N}}{\sqrt{TK}}$, then we have 
\begin{small}
	\begin{equation}\label{eq:converagerate}
		\min_{t\in[T]} \|\nabla f(x_t)\|^2 = \mathcal{O}(\frac{1}{\sqrt{NKT}}+\frac{1}{T}). 
	\end{equation}
\end{small}}
\end{corollary}

From this corollary, the convergence rate of CC-FedAvg is the same as FedAvg, which is shown in~\cite{yang2021achieving}.

\subsection{Discussion}

Compared to vanilla FedAvg, CC-FedAvg introduces two hyper-parameters $r$ and $W$ to describe clients with insufficient computational resources. 

\textbf{Influence of $r$}. 
From Corollary~\ref{cor:converagerate}, CC-FedAvg has the same order of convergence speed as FedAvg (both are $\mathcal{O}(\frac{1}{\sqrt{NKT}})$), even all the clients would skip the local training with a positive probability. 
However, $r$ affects the magnitude of the coefficients, i.e., $c$ in Eq.~(\ref{eq:nonconvex}) in Theorem~\ref{th:nonconvex}. 
From the proof of Theorem~\ref{th:nonconvex}, $c < \big(\frac{1}{2} - (6r^2 +1) L^2\frac{H_2}{K}\big)$. It can be drawn directly that a smaller $r$ is corresponding to a larger $c$, which would accelerate the convergence (note that the order of convergence is unchanged).
Therefore, $r$ has a certain influence on the convergence rate of CC-FedAvg, but is not a critical factor for convergence rate. 

\textbf{Influence of $W$}. 
Because $W$ is even not appear directly in the right side of Eq.~(\ref{eq:nonconvex}), $W$ does not directly affect the convergence rate. However, $W$ affects the range of $\eta$. When $W$ becomes larger, one of the upper bound of $\eta$ (i.e., $\frac{\sqrt{1+24r^2W^2L}-1}{12r^2KW^2L}$) gets smaller. To make Eq.~(\ref{eq:nonconvex}) hold, we should choose smaller $\eta$ for larger $W$, which leads to a larger constant term of $\mathcal{O}(\frac{1}{\sqrt{T}})$. For example, suppose $\eta$ decreases from $\frac{\sqrt{N}}{\sqrt{TK}}$ to $\frac{\sqrt{N}}{2\sqrt{TK}}$, the convergence rate is from Eq.~(\ref{eq:converagerate}) to
\begin{equation}
	\begin{small}
	min_{t\in[T]} \|\nabla f(x_t)\|^2 = \mathcal{O}(\frac{2}{\sqrt{NKT}}+\frac{1}{2\sqrt{NKT}}+\frac{1}{T}). 
\end{small}	
\end{equation}

Therefore, $W$ has a certain influence on the convergence rate of CC-FedAvg, but is not critical for convergence rate.


\section{Computation-Efficient FL}

Here we discuss a special case of CC-FedAvg with $r=1$ (denoted as CC-FedAvg($r=1$)). In this case the number of each client performing local training is decreased by $W$ compared with vanilla FedAvg. 
It indicates that we can reduce the overhead of traditional FL without sacrificing performance (Theorem~\ref{th:nonconvex} with $r=1$).
Concretely, in each training round of FL, each participant executes local training with probability $1/W$ and estimates local model with probability $(1-1/W)$ (on the contrary, each participant executes local training with probability $1$ for conventional FL). From this point of view, the field of application of CC-FedAvg is effectively broadened from addressing computation heterogeneity to any FL scenarios. It improves the computation efficiency of traditional FL without any other overhead. Exhaustive experiments are demonstrated in Section~\ref{sec:efficiency}.

\begin{table*}[t]
	\vskip -0.1in
	\centering
	\caption{\textcolor{black}{Performance (top-1 test accuracy) comparison on CIFAR-10 with different data heterogeneity, where $N=8$ and $\beta=4$. For clarity, we do not report standard deviations of all outcomes with different random seeds because they are negligibly tiny (less than 0.5).}} 
	\label{table-silo-cifar}
	\begin{center}
		\begin{small}
			\begin{tabular}{|c|c|c|c|c|c|c|c|c|c|c|c|}
				\hline
				& \multicolumn{5}{c|}{round-robin}     & \multicolumn{5}{c|}{ad-hoc}         \\ 
				\cline{2-6} \cline{7-11} 
				& Totally  & 90\%  & 80\% & 50\%  & IID   & Totally  & 90\% & 80\% & 50\% & IID \\
				& non-IID & non-IID & non-IID & non-IID  &    & non-IID & non-IID & non-IID & non-IID & \\		\hline
				FedAvg (full)    & 56.82   & 66.67   & 68.99 & 71.82&  72.88 &  56.82  & 66.67    &  68.99   & 71.82  &  72.88   \\\hdashline[1pt/1pt]
				FedAvg (dropout,best) & 40.4 & 56.98 &	59.98  & 66.87 &  67.43 & 40.4 & 56.98 &	59.98 & 66.86 & 67.43  \\\hline
				FedAvg (dropout,last) & 25.35 & 50.63 & 56.75 & 63.92  &	67.43 & 25.35 & 50.63 & 56.75 & 63.92 &  67.43 \\\hline
				Strategy 1 & 30.79   & 55.18   & 61.35   & 67.72 &69.46 &  39.45  &  60.16       &  63.95    & 70.55 &    71.95  \\\hline
				Strategy 2 & 51.45   & 61.72   & 63.37   & 67.77 &  67.35 &  48.75  & 60.18        &  61.58   & 66.73 &   66.35   \\\hline
				CC-FedAvg & \textbf{56.28}   & \textbf{65.79}   & \textbf{67.84} & \textbf{71.80} &  \textbf{72.43} &  \textbf{55.94}  &  \textbf{66.41}   & \textbf{68.61}    & \textbf{71.37}  & \textbf{72.53}   \\ 
				\hline
			\end{tabular}
		\end{small}
	\end{center}
	\vskip -0.2in
\end{table*}

\section{Experiments}

\subsection{Experimental Settings}\label{settings}
We investigate CC-FedAvg on three datasets and corresponding models as follows:
\begin{itemize}
	\item\textbf{CIFAR-10}~\cite{krizhevsky2009learning}. CIFAR-10 is made up of 60,000 32x32 color pictures which are divided into ten categories, each containing 6,000 images. The training set has 50,000 photos, whereas the test set has 10,000 images. We use a CNN network with two convolutional-pooling layers and three fully connected layers (abbr. CNN), and ResNet-18 on CIFAR-10.
	
	\item\textbf{FMNIST}~\cite{xiao2017fashion}. Fashion-MNIST is an upgraded version of MNIST. The task's complexity has risen as compared to MNIST. It contains gray-scale photos of 70,000 fashion goods from 10 categories in 28x28 dimensions. There are 7,000 photos in each category. The training set has 60000 photos, whereas the test set has 10,000 ones. 
	For FMNIST, we utilize a multi-layered perception network with 3 fully connected layers (abbr. MLP).
	
	\item\textbf{CIFAR-100}~\cite{krizhevsky2009learning}. CIFAR-100 is divided into 100 classes, each including 500 images for training and 100 images for testing. For this dateset, 
	we employ ResNet-18 with group normalization as default.
	
\end{itemize}


\begin{table*}[t]
	\centering
	\caption{\textcolor{black}{Performance (top-1 test accuracy) comparison on FMNIST with different data heterogeneity, where $N=100$ and $\beta=4$. The classes of training data are randomly distributed across clients with different computational resources.}} 
	\label{table-device-fmnist1}
	\begin{center}
		\begin{small}
				\begin{tabular}{|c|c|c|c|c|c|c|c|}
					\hline
					& 10\% & 20\% & 30\% & 40\% &  60\%  & 80\%  \\	\hline
					FedAvg (full)    & 78.76$\pm$2.62  & 80.52$\pm$1.66 & 82.44$\pm$0.73 & 82.48$\pm$0.58 & 82.71$\pm$0.30& 83.27$\pm$0.17
					\\\hdashline[1pt/1pt]
					FedAvg (dropout,best) &74.50$\pm$3.10 &	77.76$\pm$3.28 &	78.19$\pm$2.92 & 79.48$\pm$2.71 & 81.58$\pm$0.51 & 82.26$\pm$0.34\\\hline
					FedAvg (dropout,last) & 57.43$\pm$10.06 & 67.38$\pm$8.76	& 73.72$\pm$3.48 & 75.74$\pm$2.30 &78.24$\pm$0.40 & 79.24$\pm$0.24 \\\hline
					Strategy 1 & 60.12$\pm$12.51 & 74.61$\pm$4.79 & 78.90$\pm$2.65 & 80.61$\pm$0.82 & 80.47$\pm$0.71& 81.35$\pm$0.39 \\\hline
					Strategy 2 & 66.58$\pm$3.66 & 73.63$\pm$3.86 & 77.76$\pm$1.61 & 78.91$\pm$0.70  & 80.12$\pm$0.30& 81.52$\pm$0.26 \\\hline
					CC-FedAvg & \textbf{75.78$\pm$2.30} & \textbf{78.40$\pm$2.50} &\textbf{81.70$\pm$0.81} & \textbf{81.42$\pm$0.31} & \textbf{82.46$\pm$0.63} & \textbf{83.23$\pm$0.26} \\ 
					\hline
				\end{tabular}
		\end{small}
	\end{center}
	\vskip -0.1in
\end{table*}

\begin{figure*}[ht]
	\centering
	\subfloat[ResNet-18 on CIFAR-10]{
		\includegraphics[width=0.6\columnwidth]{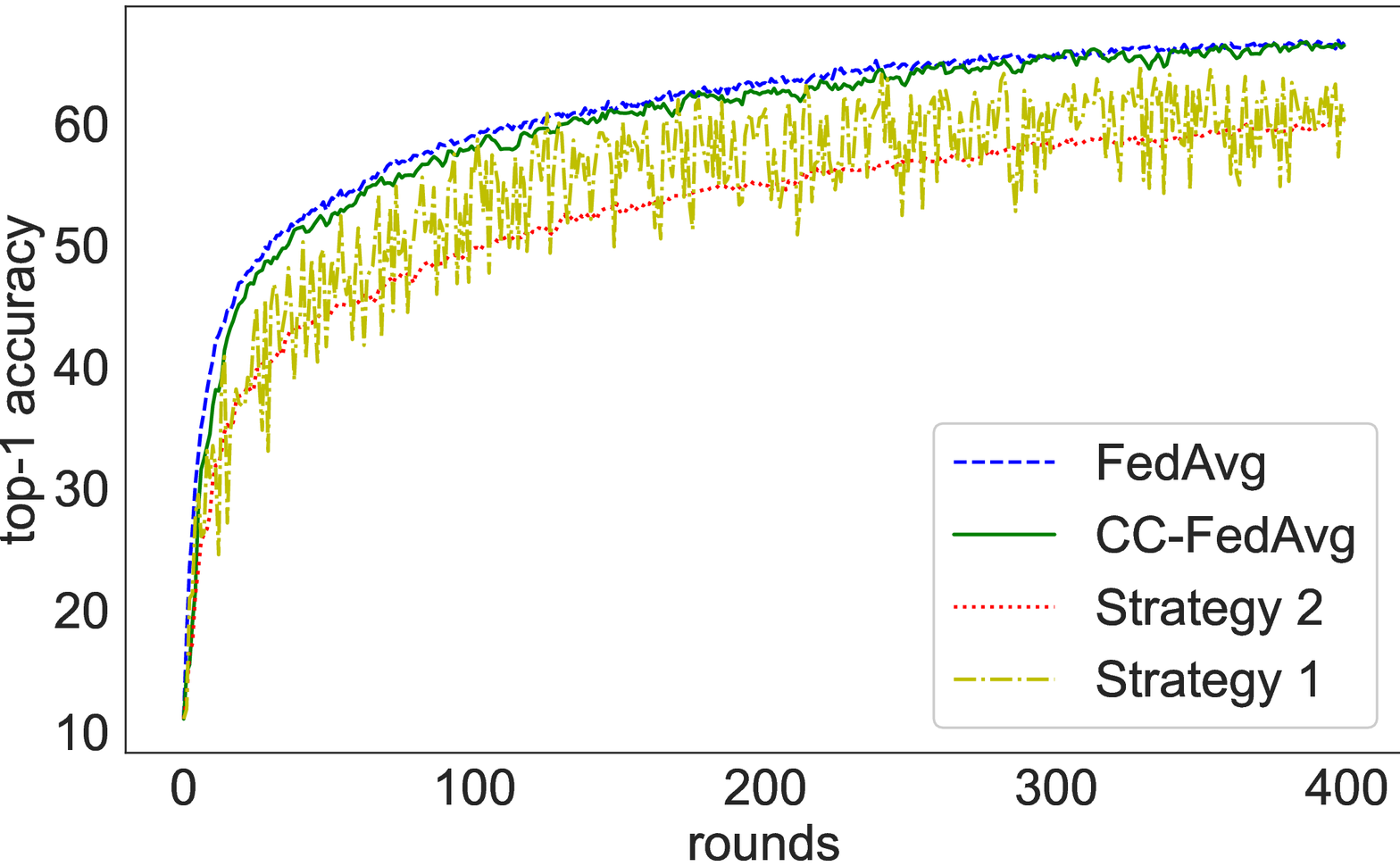} 
		\label{fig:cifar10} 
	}
\hfill
	\subfloat[MLP on FMNIST]{
		\includegraphics[width=0.6\columnwidth]{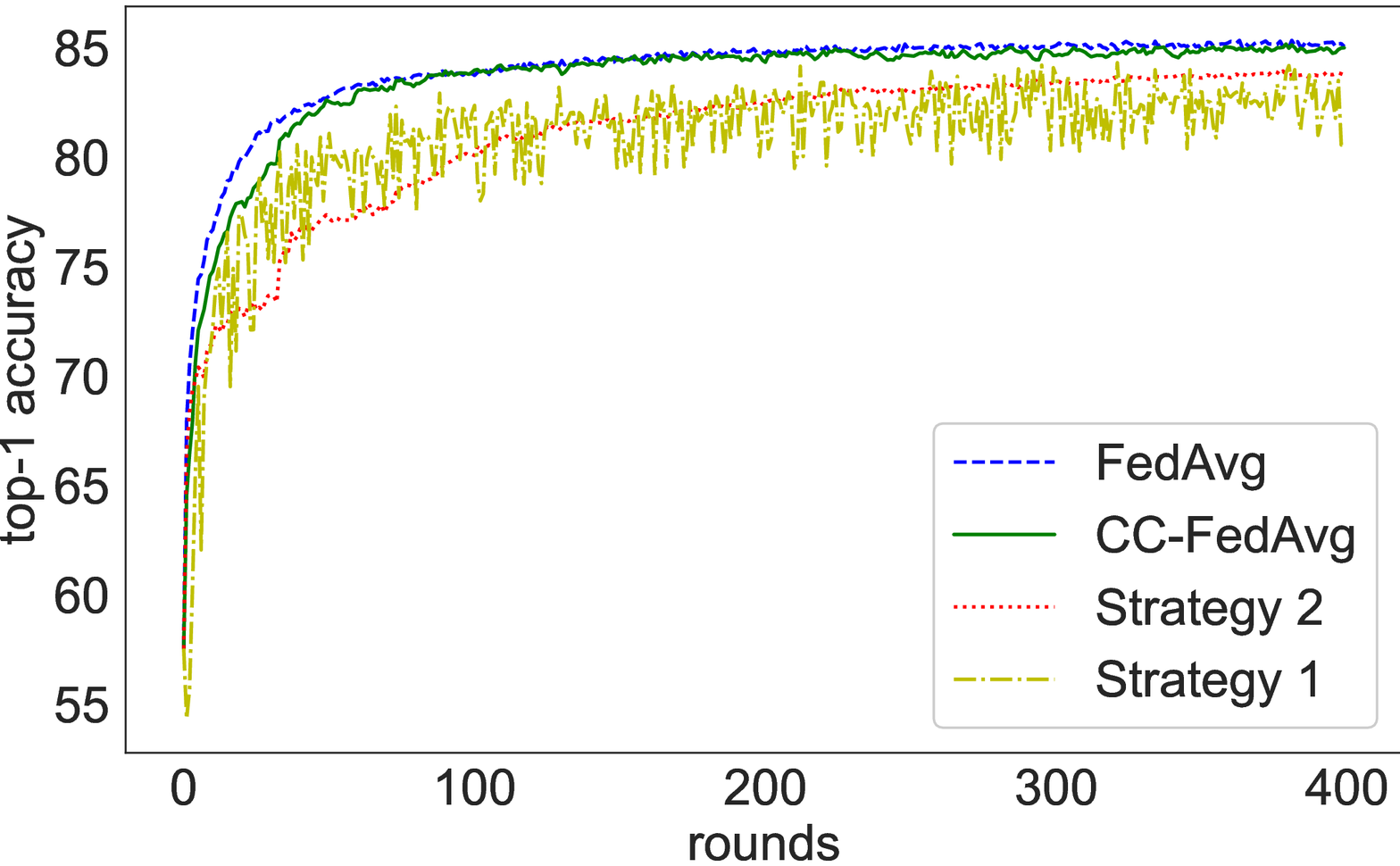} 
		\label{fig:fmnist} 
	}
\hfill
	\subfloat[ResNet-18 on CIFAR-100]{
		\includegraphics[width=0.6\columnwidth]{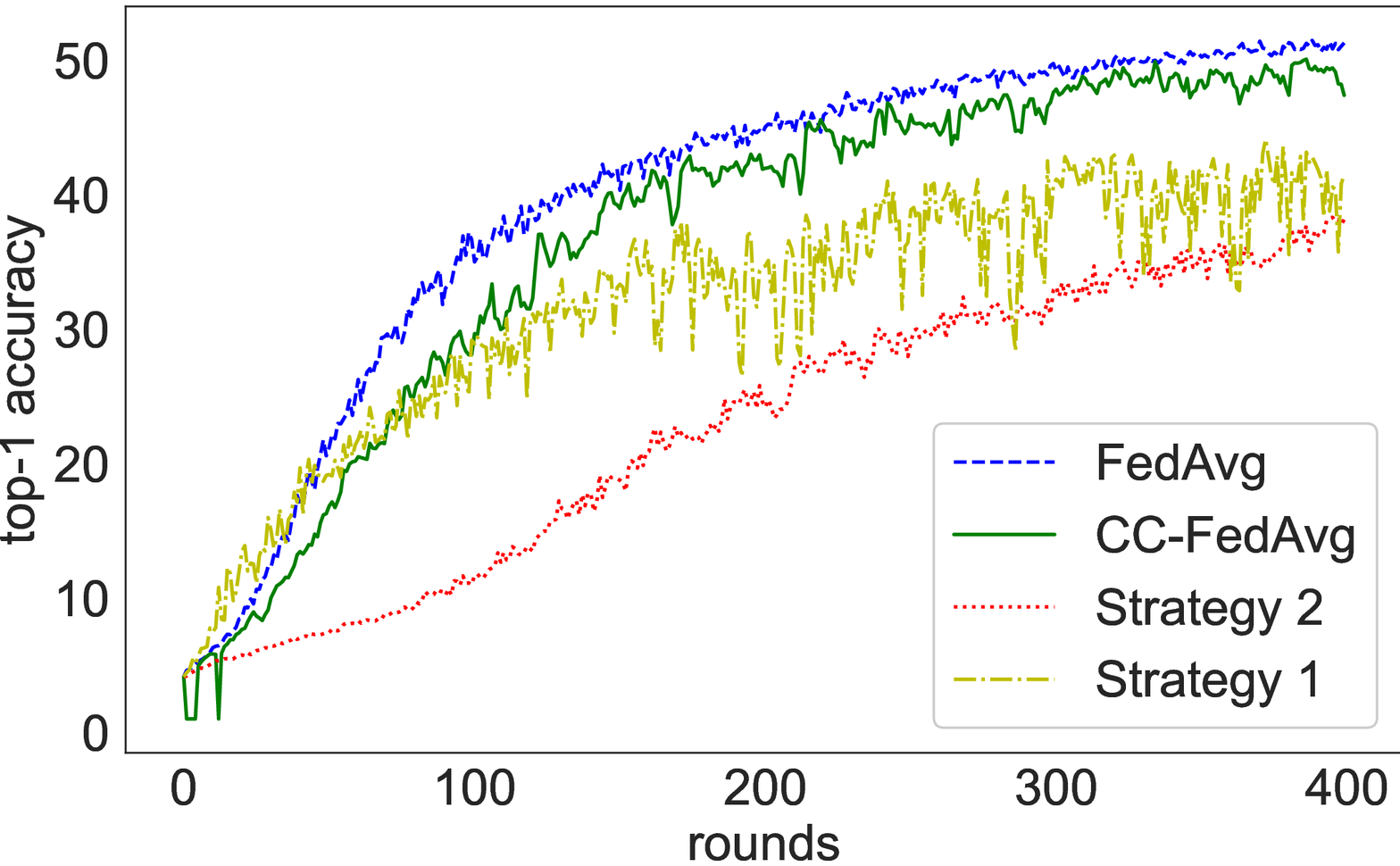} 
		\label{fig:cifar100} 
		
	}
	\caption{Performance comparison on multiple datasets.}
	\label{fig:convergence} 
\end{figure*}

We evaluate the proposed CC-FedAvg on tasks involving the customization of computational resources. To facilitate our discussion, we assume a specific heterogeneous training budget with an imbalanced resource distribution among clients. Specifically, for client $i$, $p_i=(1/2)^{\lfloor \frac{\beta \cdot i}{N}\rfloor}$, where $\beta$ controls the number of heterogeneous resource levels. In this situation, $r$ is about $1-\frac{1}{\beta}$, because there are $\lceil \frac{N}{\beta} \rceil$ clients with $p_i=1$. For instance, if we set $\beta=3$, all the clients are equally divided into 3 groups in which the clients participate in 1, 0.5 and 0.25 times of the expected number of training round, respectively. Concretely, for a cross-silo scenario with 30 clients, if it takes 400 rounds for training, each group of 10 clients performs local training in 400, 200, 100 rounds, respectively. 
We construct data heterogeneity following~\cite{zhang2022fedcos}, using $\gamma$ to control the degree of data heterogeneity (the proportion of non-IID data local data across clients). Particularly, $\gamma=0$ means the data cross clients is totally non-IID (named totally non-IID), and $\gamma=1$ means the data is IID cross clients (named IID).

For the client with $p_i < 1$, we consider two types of scheduling methods to save the computational resources: 

\textbf{``round-robin'' schedule}: Client $i$ skips $(1/p_i-1)$ rounds after performing local training once strictly. This schedule simulates the situation that the client can schedule tasks in advance. For example, if the client has insufficient energy, it can plan the rounds ahead of time to guarantee that it has enough energy to endure until the end of training. 

\textbf{``ad-hoc'' schedule}: Client $i$ skips each round with probability $1-p_i$. This schedule simulates the situation that the client can schedule tasks in time. For example, if the computational resources of clients vary randomly throughout training due to the addition and removal of other irrelevant tasks, the clients can decide whether to perform local training based on real-time computational resources. 


\begin{figure*}[ht]
	\centering
	\subfloat[CNN on CIFAR-10]{
		\includegraphics[width=0.62\columnwidth]{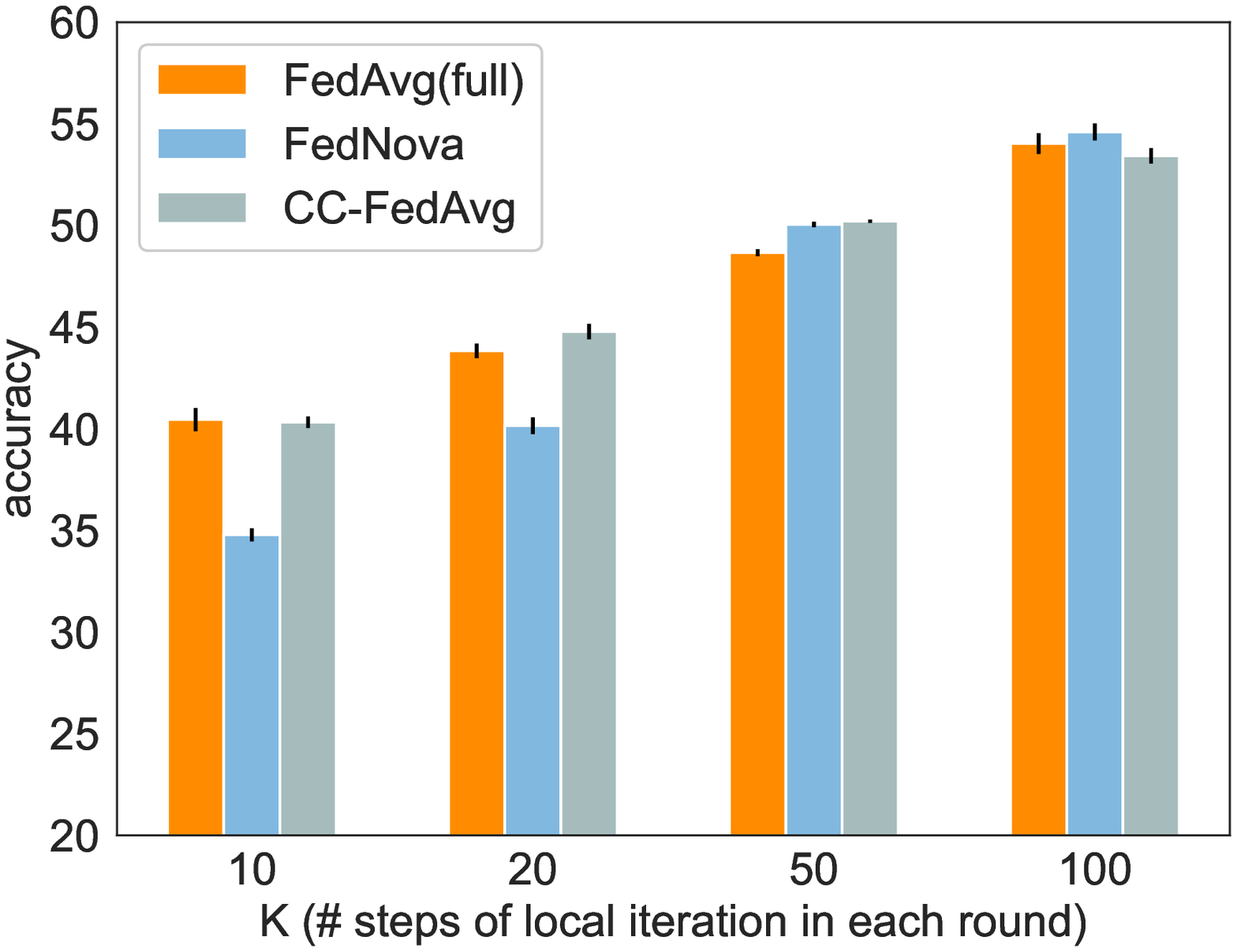} 
		\label{fig:nova_cifar10} 
	}
	\quad 
	\subfloat[ResNet-18 on CIFAR-100]{
		\includegraphics[width=0.62\columnwidth]{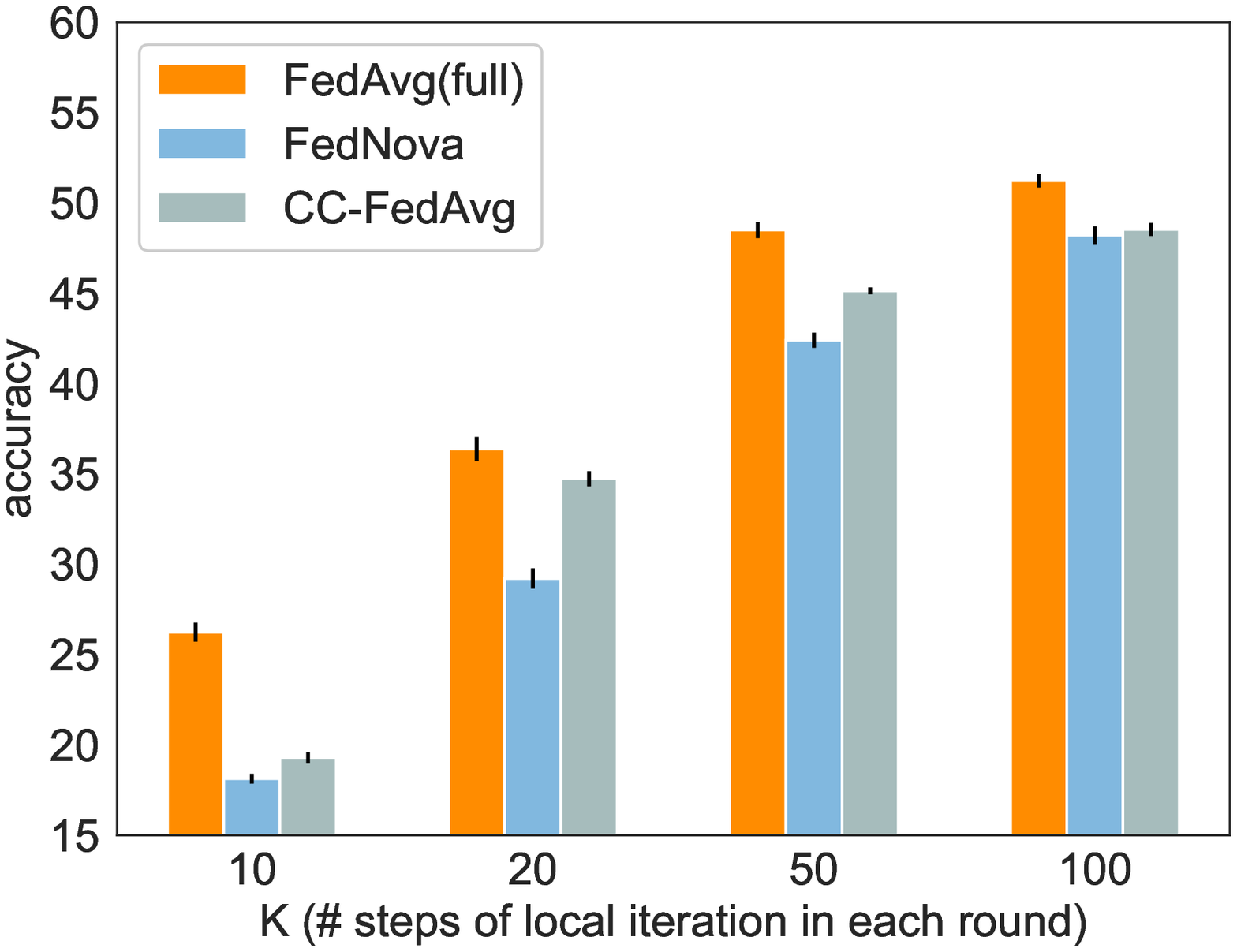} 
		\label{fig:nova_cifar100} 
	}
	\quad 
	\subfloat[\textcolor{black}{performance comparison with different T (K=10, CNN on CIFAR-10)}]{
		\includegraphics[width=0.62\columnwidth]{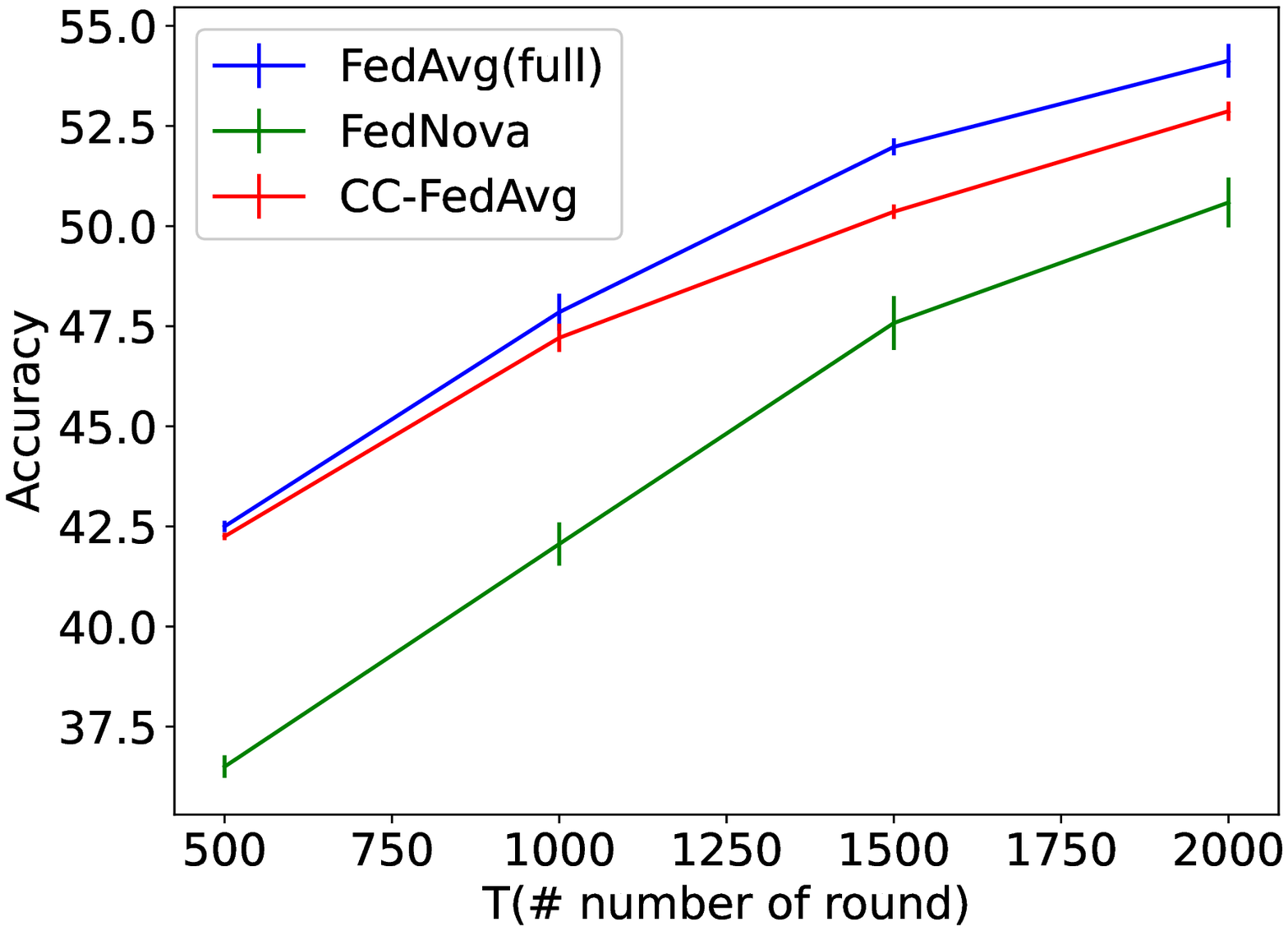} 
		\label{fig:nova_local10} 
	}
	\caption{Performance comparison between CC-FedAvg and FedNova.}
	\label{fig:nova} 
\end{figure*}

\subsection{Performance}
Table~\ref{table-silo-cifar} compares the performance of CC-FedAvg, FedAvg and other two strategies in cross-silo setting. 
We consider three kinds of FedAvg. FedAvg (full) is FedAvg without the limitation of computational resources that all clients participate in training every round. FedAvg (dropout) denotes FedAvg with computation limitation (due to energy deficiency) that the clients would drop out after the quota of participation round is exhausted. For instance, if total number of round is 400, the clients with $p_i=0.5$ would drop out after 200 rounds of training. FedAvg (dropout, best) and FedAvg (dropout, last) indicates the best and the last performance of the global model, respectively. 
We construct a variety of federated scenarios containing 8 clients on CIFAR-10 with different data heterogeneity. 
We use CNN model, and set SGD as local optimizer with learning rate 0.01. All methods perform 400 rounds, with 3 epochs in each round. To simulate heterogeneous computational resources, we set $\beta=4$, i.e., the clients are on 4 different levels based on the computational resource budgets, performing local training with probabilities 1, 0.5, 0.25, 0.125, respectively. We investigate the model performance on both ``round-robin'' schedule and ``ad-hoc'' schedule. The result shows that CC-FedAvg outperforms FedAvg (dropout) and the other two strategies regardless of data heterogeneity and task schedules. Remarkably, compared to FedAvg (full), 
CC-FedAvg achieves comparable performance, even though 75\% clients are not fully involved, and 25\% clients participate in only 1/8 of the training rounds. Fig.~\ref{fig:8worker} in Appendix~\ref{visualization} visualizes the participation of the clients. Compared with FedAvg, CC-FedAvg requires much less computational resources.
Furthermore, comparing the performance between two schedules in Table~\ref{table-silo-cifar}, CC-FedAvg performs similarly regardless data distribution. Without loss of generality, in the following we set ``ad-hoc'' schedule as default.


%
%
%

To compare the performance of CC-FedAvg and other two strategies in cross-device setting, we construct federated scenarios on FMNIST with 100 clients where each client contains two classes of training data, and varying participant ratios decided by the server (from 10\% to 80\%). We specify $\beta=4$. All methods run 400 rounds with 200 iterations in each round. 
To better understand the participation, we visualize the client participation of training in Fig.~\ref{fig:100client} of Appendix~\ref{visualization}. 
The results are shown in Table~\ref{table-device-fmnist1}, 
where CC-FedAvg performs comparable to FedAvg, with a performance decline of no more than 3\%, and consistently outperforms other two strategies and FedAvg (dropout) with energy limitation. 
Furthermore, to study the fluctuations of model performance, we perform each method 5 times with various random seeds. For all methods, the standard deviation decreases as the proportion of participants increases.
CC-FedAvg performs as consistently as FedAvg, where the standard deviation is less than 3.0 for all participant ratios. On the contrary, the performances of models trained by other two methods varies considerably, especially with low participant ratios (e.g., 10\%).

\subsection{Convergence}

We compare the convergence curves of CC-FedAvg and FedAvg as well as other two baselines on three datasets in Fig.~\ref{fig:convergence}. We construct the scenarios with 8 clients and $\beta=4$ under 90\% non-IID data setting. CC-FedAvg exhibits convergence curves that are almost identical to FedAvg, especially on two simpler datasets (FMNIST and CIFAR-10). 
It means that with CC-FedAvg, there is essentially no influence from computational underpower throughout the model training process, even 25\% participants only perform 1/8 number of iterations compared with them in FedAvg. 
It also visually verifies Corollary~\ref{cor:converagerate} that CC-FedAvg has the same convergence rate as FedAvg.
On the contrary, \textcolor{black}{Strategy 1 performs very unstable that the curves wobble a lot due to the bias aggregation in each round.} Strategy 2 has a stable convergence curve, but it is significantly lower than FedAvg and CC-FedAvg.

\begin{figure*}[ht]
	\centering
	\subfloat[cross-silo scenarios (CNN on CIFAR-10)]{
		\includegraphics[width=0.85\columnwidth]{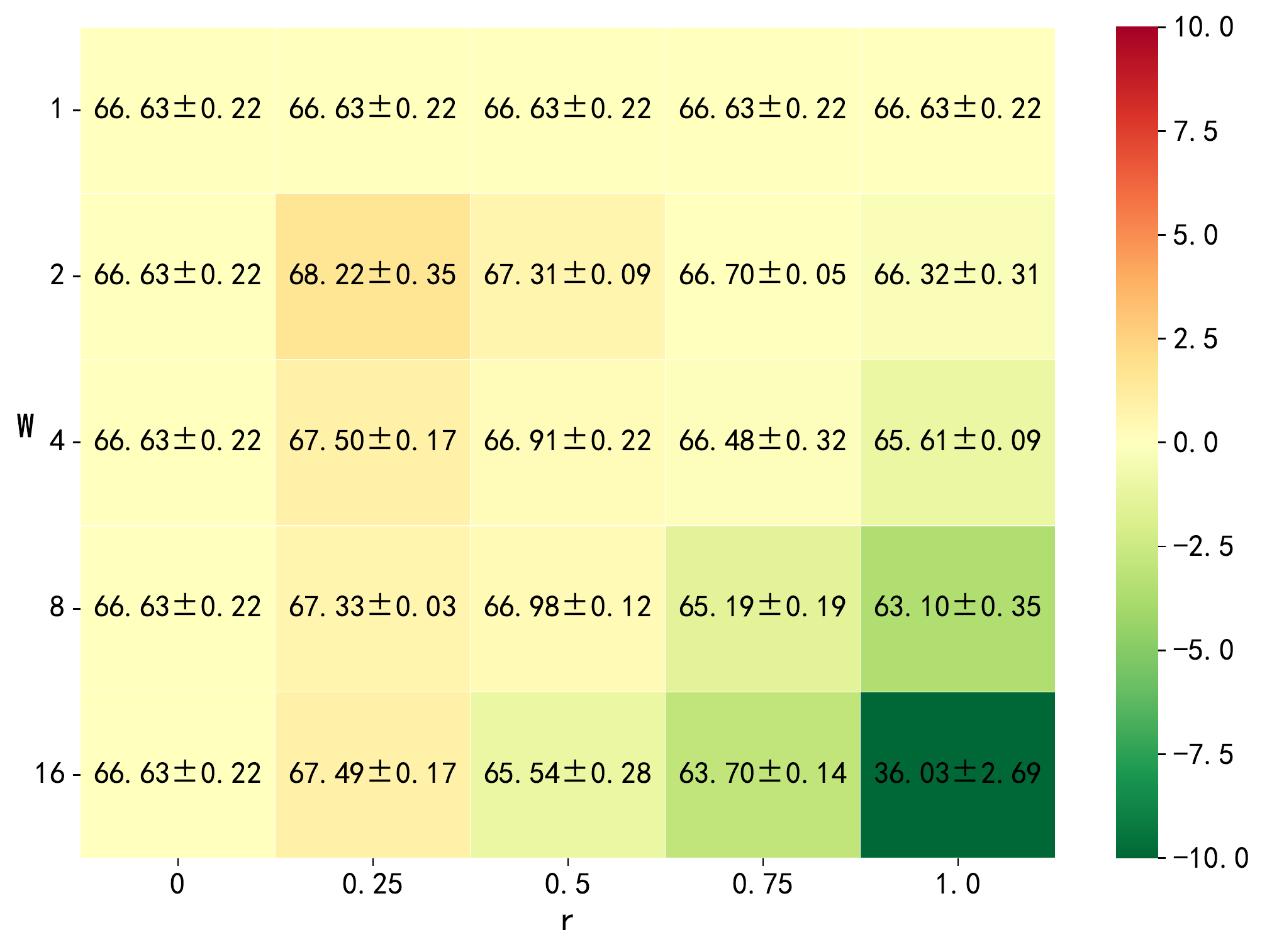} 
		\label{fig:heatmap_cifar10_silo_md4} 
	}
	\quad 
	\subfloat[cross-device scenarios (MLP on FMNIST)]{
		\includegraphics[width=0.85\columnwidth]{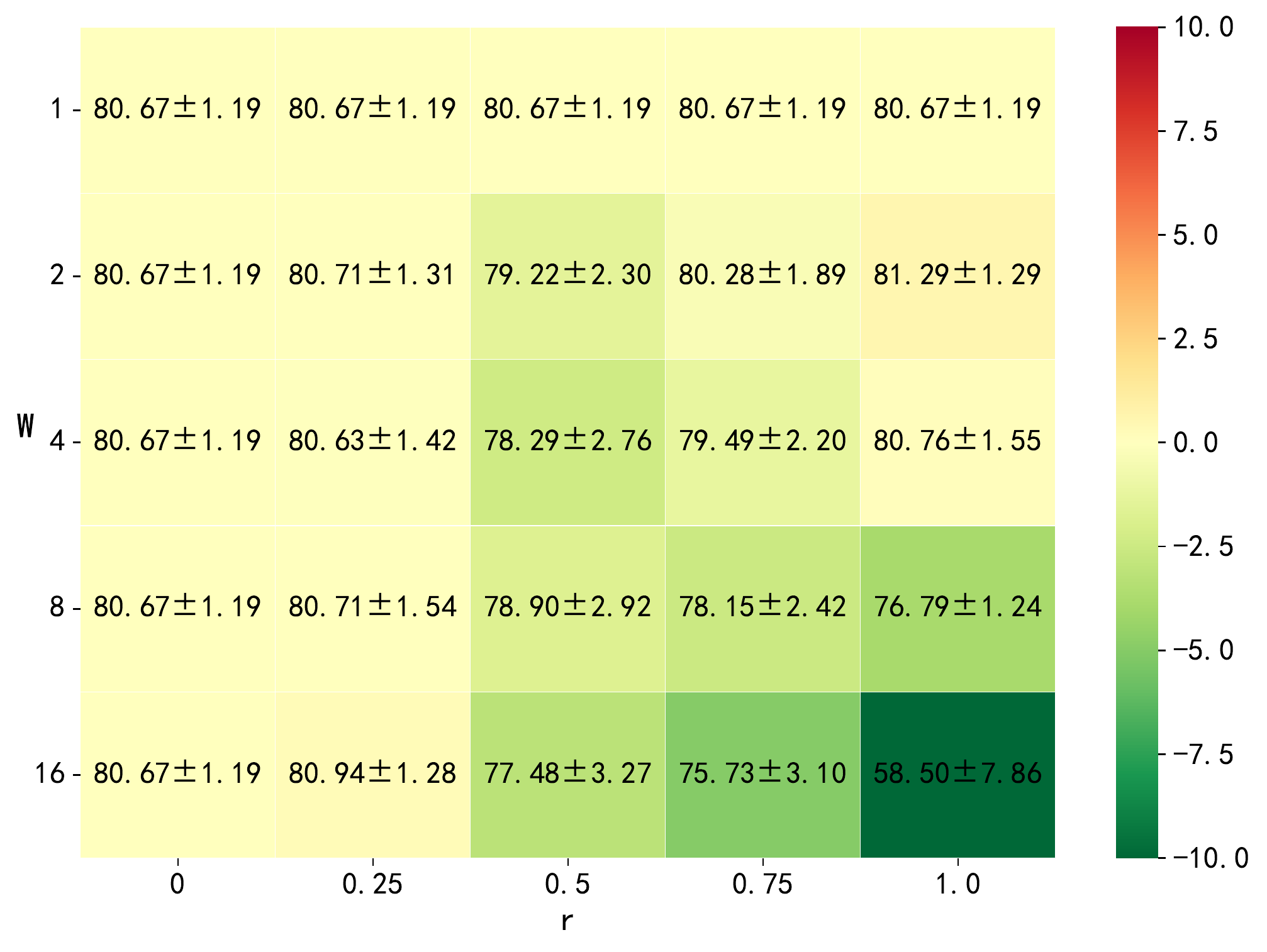} 
		\label{fig:heatmap_fmnist_device_md4} 
	}
	\caption{Performance changes with varying of $r$ and $W$ (CC-FedAvg). The color at each grid indicates the performance difference between the model of CC-FedAvg with corresponding $r$ and $W$ and FedAvg (i.e., CC-FedAvg with $r=0$ or $W=0$). The number at each grid indicates the top-1 accuracy on test dataset.}
	\label{fig:MandW} 
	\vskip -0.2in
\end{figure*}

\begin{figure}[htbp]
	\centering
	\subfloat[cross-silo scenarios (CNN on CIFAR-10)]{
		\includegraphics[width=0.7\columnwidth]{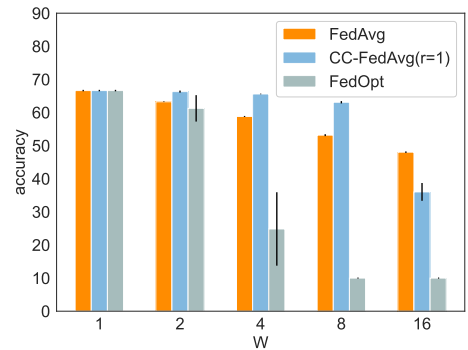} 
		\label{fig:efficient_cifar10_2} 
	}
	\quad 
	\subfloat[cross-device scenarios (MLP on FMNIST)]{
		\includegraphics[width=0.7\columnwidth]{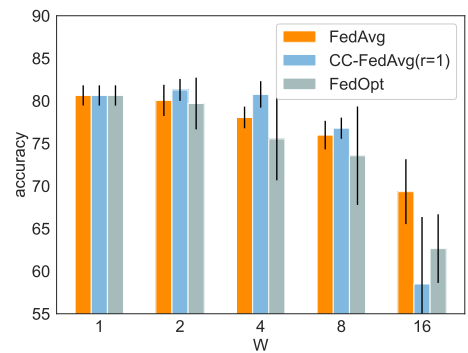} 
		\label{fig:efficient_fmnist_2} 
	}
	\caption{Efficiency comparison between CC-FedAvg($r=1$) with FedAvg.}
	\label{fig:efficient} 
\end{figure}

\subsection{Comparison with FedNova}\label{sec:nova}

FedNova~\cite{wang2020tackling} can be used self-adapt the computation resources of clients. It also can be regarded as a strategy of model estimation, using the model with inadequate training. As a result, FedNova performs well depend on the adequacy of local training. In Fig.~\ref{fig:nova} we compare the performance of FedNova and CC-FedAvg. In Fig.~\ref{fig:nova_cifar10}, we construct the comparison under totally non-IID data distribution following the setting of Table~\ref{table-silo-cifar}. We turn the number of local iteration in each round $K$ from 10 to 100. It illustrates that the performance of CC-FedAvg is stable, which is slightly lower than FedAvg (full). On the contrary, FedNova performs much worse than others at $K=10$, and it becomes better when $K$ increases. It is because local estimation is not accurate when $K$ is small. For example, when $K=10$, the clients with $p_i=1/8$ would perform local iterations at most 2 times. When $K$ is large, local models are good enough even only $K/8$ iterations are performed. Actually, it is effected by the hardness of task and the complexity of model. In Fig.~\ref{fig:nova_cifar100} we illustrate the comparison by ResNet-18 on CIFAR-100, following the set of~Fig.~\ref{fig:cifar100}. It shows the similar results as Fig.~\ref{fig:nova_cifar10}. Furthermore, due to the complexity of model and task, larger number of local iterations is required to obtain a good estimation by FedNova. Compared with Fig.~\ref{fig:nova_cifar10}, where FedNova performs as good as others when $K$ is larger than 50, in Fig.~\ref{fig:nova_cifar100} the corresponding $K$ is larger than 200.  Furthermore, we find the weakness of FedNova for small $K$ cannot be made up by longer training. In Fig.~\ref{fig:nova_local10} we extend the total number of training rounds up to 2000 for $K=10$ in Fig.~\ref{fig:nova_cifar10}. Although the performance of FedNova improves as the number of training rounds increases, the performance gap with other methods remains unchanged. Therefore, FedNova can be applied in FL with computation heterogeneity in limited scenarios. 

\subsection{Factors of CC-FedAvg}

To understand the effect of key factors of CC-FedAvg, namely $r$ and $W$, we construct federated scenarios by varying these two hyper-parameters. Participants are divided into two groups, where the first group contains $1-r$ of participants with sufficient computational resources, and the second group consists of the rest part of participants with the same $W$ (the corresponding $p_i=\frac{1}{W}$). 
Particularly,  CC-FedAvg is degraded into FedAvg when $W=1$ or $r=0$.

Fig.~\ref{fig:MandW} demonstrates the model performance by gridding the values of $r$ and $W$, where Fig.~\ref{fig:heatmap_cifar10_silo_md4} illustrates the performances in cross-silo scenarios, and Fig.~\ref{fig:heatmap_fmnist_device_md4} is in cross-device scenarios, where only 20\% clients are selected by the server in each round. With the increment of $W$ or $r$, the performance is essentially steady, unless both $r$ and $W$ are extremely large. 
In Fig.~\ref{fig:heatmap_cifar10_silo_md4}, CC-FedAvg even surpasses FedAvg while requiring less computing effort (e.g., $r=0.25$). It indicates that excessive computing resources (full participation in each round) are not necessarily beneficial to accelerate the training. 
In both figures, when $r=1.0$ and $W=16$ the model performance of CC-FedAvg degrades significantly even more than that of Strategy~\ref{strategy1} and Strategy~\ref{strategy2} (the details are in Fig.~\ref{fig:MandW_cifar10} and Fig.~\ref{fig:MandW_fmnist} in Appendix~\ref{sub:influence}). In this situation, each client executes local training only once about every 16 rounds. The uploaded local models in most rounds are quite inaccurate since they are guessed based on outdated information. When $r$ or $W$ are reduced, the performance of CC-FedAvg rapidly improves since more accurate information is introduced.

\subsection{Computational Efficiency Improvement}\label{sec:efficiency}

We compare the performance of CC-FedAvg($r=1$) with FedAvg under the same computation overhead. Specifically, for each participant in CC-FedAvg($r=1$), the number of rounds to perform local training is only $1/W$ of corresponding FedAvg. Therefore, to compare both methods with the same computation resources budgets, we compare the performance of CC-FedAvg($r=1$) (and the number of training round $T$) and FedAvg with total training round $T/W$. We demonstrate the comparison both under cross-silo (as Fig.~\ref{fig:heatmap_cifar10_silo_md4}) and cross-device (as Fig.~\ref{fig:heatmap_fmnist_device_md4}) settings. When $W$ is moderate large (e.g., $W\leq 8$), CC-FedAvg($r=1$) outperforms FedAvg. However, when if $W$ is too large (e.g., $W = 8$), CC-FedAvg underperforms FedAvg, because the local estimations are far from the true models. 
Therefore, by setting proper $W$, CC-FedAvg($r=1$) can be an alternative with more computational efficiency to FedAvg.

For CC-FedAvg($r=1$) in Fig.~\ref{fig:efficient} each client skips or performs local training in ad-hoc schedule individually. In each round, there will always be clients undertaking local training to keep true information in aggregation. If we synchronize the rhythm of all clients that skipping or performing training in all rounds, CC-FedAvg becomes similar as FedOpt~\cite{reddi2021adaptive}. 
For example, if all clients estimate local models in sequential $W-1$ rounds after execute local training in one round (denote as round $t$), the aggregated model at round $t+W-1$ are $x_t+(W-1)\Delta_t$, where $(W-1)$ can be viewed as the global learning rate in FedOpt. \textcolor{black}{In Fig.~\ref{fig:efficient} we also add the comparison between FedOpt and CC-FedAvg($r=1$). It shows that FedOpt performs much worse than CC-FedAvg($r=1$), even worse than FedAvg in this situation. Therefore, the ad-hoc schedule plays a vital role to keep CC-FedAvg's performance.}

\begin{table}[tbp]
	\centering
	\caption{\textcolor{black}{Performance Comparison of different local model estimation strategy on FMNIST and CIFAR-10.}} 
	\label{tb:compare_combination}
	\vskip 0.15in
	\begin{tabular}{|c|c|c|c|c|}
		\hline
		&                                                           & Strategy 2    & CC-FedAvg & CC-FedAvg(c) \\ \hline
		\multicolumn{1}{|c|}{\multirow{3}{*}{CIFAR-10}} & \begin{tabular}[c]{@{}l@{}}Totally\\ non-IID\end{tabular} & 48.76$\pm$0.06         & \textbf{56.03$\pm$0.08}     &  53.15$\pm$0.11            \\ \cline{2-5} 
		\multicolumn{1}{|c|}{}                          & \begin{tabular}[c]{@{}l@{}}80\%\\ non-IID\end{tabular}    & 61.54$\pm$0.11         & \textbf{68.59$\pm$0.20}     &   64.45$\pm$0.16           \\ \cline{2-5} 
		\multicolumn{1}{|c|}{}                          & \begin{tabular}[c]{@{}l@{}}50\%\\ non-IID\end{tabular}    & 66.68$\pm$0.13         & \textbf{71.34$\pm$0.02}     & 69.11$\pm$0.12             \\ \hline
		\multirow{3}{*}{FMNIST}                         & 20\%                                                      & 73.6$\pm$3.86 & 78.40$\pm$2.50     & \textbf{79.07$\pm$1.34}        \\ \cline{2-5} 
		& 40\%                                                      & 78.91$\pm$0.70     & 81.42$\pm$0.31     & \textbf{82.02$\pm$0.37}         \\ \cline{2-5} 
		& 60\%                                                      & 80.12$\pm$0.30         & \textbf{82.46$\pm$0.63}     & 82.05$\pm$0.10     \\ \hline
	\end{tabular}
\end{table}

\subsection{\textcolor{black}{Replacement of local model Estimation}}\label{sec:estimation}
\textcolor{black}{Except Strategy~\ref{strategy3}, CC-FedAvg can embed other local model estimations. Generally, Strategy~\ref{strategy2} can be regarded as an alternative method but the effect is not as good as Strategy~\ref{strategy3}. In section~\ref{sec:method} we mention that Eq.~(\ref{conbine}) is a replacement to construct a new CC-FedAvg. In order to distinguish this method and the default one, we call the new one as CC-FedAvg(c). Table~\ref{tb:compare_combination} show the performance comparison, where the settings are the same as Table~\ref{table-silo-cifar} on CIFAR-10 (``cross-silo'' scenario with alterable data heterogeneity) and Table~\ref{table-device-fmnist1} on FMNIST (``cross-device'' scenario with different participant ratios). The threshold $\tau$ is 100. It shows that CC-FedAvg(c) also is an acceptable method that outperforms Strategy~\ref{strategy2} persistently, even is better than the default CC-FedAvg in some cases. Therefore, we can choose the appropriate strategy of local model estimation for given tasks to obtain better models.}  

\section{Conclusion}
In this paper, we present a strategy for estimating local models with little overhead. Based on it, we propose a novel federated learning method CC-FedAvg for customizing computational resources to address the challenge of computation heterogeneity. Both theoretical analysis and extensive experiments demonstrate the effectiveness of CC-FedAvg. Meanwhile, CC-FedAvg can be viewed as a computation-efficient extension of FedAvg that it can replace FedAvg in any scenarios without be limited to tackle computation heterogeneity.

\appendices

\section{Variants of CC-FedAvg}\label{sec:variantsCC-FedAvg}

\begin{algorithm}[htb]
	\caption{\textbf{CC-FedAvg}: Computationally Customized Federated Averaging (historical information backup on the server)} 
	\label{al:ccfl2}
	\begin{algorithmic}[1]
		\STATE Initialize $x_0$.
		\FOR {$t = 0$ to $T-1$}
		\STATE Server selects $\mathcal{S}_t$ randomly and sends $x_{t}$ to clients in $\mathcal{S}_t$.
		\STATE (At client:)
		\FOR {each client $i \in \mathcal{S}_t$ in parallel}
		\IF {not skip this round} 
		\STATE /* with probability $p_i$ */
		\STATE initial local model: $x_{t,0}^i = x_{t}$.  \label{line:ltstart2}
		\FOR {$k=0$ to $K-1$}
		\STATE Update local model by an unbiased estimate of gradient: $x^i_{t,k+1} = x^i_{t,k} - \eta g_{t,k}^i$.
		\ENDFOR
		\STATE Get $\Delta^i_t = x^i_{t,K} - x^i_{t,0}$.\label{line:ltend2}
		\STATE Send $\Delta^i_t$ back to the server.
		\ELSE  
		\STATE /* with probability $1-p_i$ */
		\STATE Send ($i$,``skip'' signal) to the server.
		\ENDIF
		\ENDFOR
		\STATE (At server:)
		\IF {Receive ($i$,``skip'' signal)}
		\STATE Get $\Delta^i_t = \Delta^i_{t-1}$.\label{line:skip2}
		\ENDIF
		\STATE $\Delta_t = \frac{1}{|\mathcal{S}_t|}\sum_{i \in \mathcal{S}_t} \Delta^i_t$.
		\STATE Update global model $x_{t+1} = x_t + \Delta_t$.
		\ENDFOR
	\end{algorithmic}
\end{algorithm}

\begin{algorithm}[htb]
	\caption{\textbf{CC-FedAvg}: Computationally Customized Federated Averaging (mixed backup of historical information)} 
	\label{al:ccfl3}
	\begin{algorithmic}[1]
		\STATE Initialize $x_0$.
		\FOR {$t = 0$ to $T-1$}
		\STATE Server selects $\mathcal{S}_t$ randomly and sends $x_{t}$ to clients in $\mathcal{S}_t$.
		\STATE (At client:)
		\FOR {each client $i \in \mathcal{S}_t$ in parallel}
		\IF {not skip this round} 
		\STATE /* with probability $p_i$ */
		\STATE initial local model: $x_{t,0}^i = x_{t}$. \label{line:ltstart3}
		\FOR {$k=0$ to $K-1$}
		\STATE Update local model by an unbiased estimate of gradient: $x^i_{t,k+1} = x^i_{t,k} - \eta g_{t,k}^i$.
		\ENDFOR
		\STATE Get $\Delta^i_t = x^i_{t,K} - x^i_{t,0}$.\label{line:ltend3}
		\STATE Send $\Delta^i_t$ back to the server.
		\ELSE  
		\STATE /* with probability $1-p_i$ */
		\IF {$\Delta^i_{t-1}$ is backed up locally}
		\STATE Get $\Delta^i_t = \Delta^i_{t-1}$.\label{line:skip3}
		\STATE Send $\Delta^i_t$ back to the server.
		\ELSE
		\STATE Send ($i$,``skip'' signal) to the server.
		\ENDIF
		\ENDIF
		\ENDFOR
		\STATE (At server:)
		\IF {Receive ($i$,``skip'' signal)}
		\STATE Get $\Delta^i_t = \Delta^i_{t-1}$.\label{line:skip31}
		\ENDIF
		\STATE $\Delta_t = \frac{1}{|\mathcal{S}_t|}\sum_{i \in \mathcal{S}_t} \Delta^i_t$.
		\STATE Update global model $x_{t+1} = x_t + \Delta_t$.
		\ENDFOR
	\end{algorithmic}
\end{algorithm}

In Algorithm~\ref{al:ccfl}, each client saves $\Delta_i^t$ for further use after performing 
line~15, which leads to extra storage cost. Actually, since the server knows $\Delta_{t-1}^i$ if the historical information is saved, the server does not need the client to return the estimated movement of the local model. Instead, the server can perform 
line~15 directly when it knows the corresponding client would skip this round (for example, the client returns a ``skip'' signal if it would skip the round). In this situation, the overhead of clients is further reduced: besides reducing the computational overhead, both the storage overhead and communication overhead are reduced. Specifically, the storage of $\Delta_{t-1}^i$ is transferred to the server. The amount of data being transmitted is reduced from thousands of even millions of bytes (determined by the number of model parameters) to at least 1 bit (indicating ``skip'' or not). We propose another variant of CC-FedAvg (Algorithm~\ref{al:ccfl2}) without increasing the storage overhead of clients. Furthermore, the two schemes that the historical information is stored locally or on the server can be mixed to obtain the third variant of CC-FedAvg (Algorithm~\ref{al:ccfl3}). For the client with sufficient storage and good communication conditions, the historical information can be saved by the client itself and performs CC-FedAvg as Algorithm~\ref{al:ccfl}, otherwise the historical information is saved by the server. By this hybrid scheme, we can balance the local requirement and the server load. After all, it is also a great burden to the server if all the local historical information is stored on the server.

\section{Important Lemmas}

To understand Theorem~\ref{th:nonconvex} better, we list some important Lemmas here.

\begin{lemma}\label{lemma:movementsum}
	\textit{For learning rate $\eta \leq \frac{1}{4LK}$, we have 
	\begin{equation}
		\begin{aligned}
		&\sum_{k=0}^{K-1}\mathbb{E}\| x_{t,k}^i - x_t\|^2 \\
		\leq& 5K^2\eta^2\sigma_L^2+30K^3\eta^2\sigma_G^2+30K^3\eta^2\|\nabla f(x_t)\|^2.			
		\end{aligned}
	\end{equation}}
\end{lemma}

\begin{IEEEproof}	
	\begin{align*}\label{eq:multiterm}
			&\mathbb{E}\| x_{t,k}^i - x_t\|^2 = \mathbb{E}\| x_{t,k-1}^i - \eta g_{t,k-1}^i - x_t\|^2 \\
			\leq& \mathbb{E}\|\underbrace{x_{t,k-1}^i - x_t}_{a} -\eta (\underbrace{g_{t,k-1}^i-\nabla f_i(x_{t,k-1}^i)}_{b} \\
			& \  + \underbrace{\nabla f_i(x_{t,k-1}^i) - \nabla f_i(x_{t})}_{c} \\
			& \  + \underbrace{\nabla f_i(x_{t}) - \nabla f(x_{t})}_{d} + \underbrace{\nabla f(x_{t})}_{e})\|^2  \\
			\overset{(a_1)}{=}& \mathbb{E}\|a -\eta (c + d + e)\|^2 + \eta^2\mathbb{E}\|b\|^2 \\
			\overset{(a_2)}{\leq}& (1+ \frac{1}{2K-1})\eta^2\mathbb{E}\|a\|^2 + \eta^2\mathbb{E}\|b\|^2 \\
			&\  + 2\eta^2 K(3\mathbb{E}\|c\|^2 + 3\mathbb{E}\|d\|^2 + 3\mathbb{E}\|e\|^2) \\
			\overset{(a_3)}{\leq}& (1+ \frac{1}{2K-1})\eta^2 \mathbb{E}\|x_{t,k-1}^i - x_t\|^2 + \eta^2\sigma_L^2 \\
			&\  + 6KL^2\eta^2\mathbb{E}\|x_{t,k-1}^i-x_{t}\|^2+ 6K\eta^2\sigma_G^2 \\
			&\   + 6K\eta^2\| \nabla f(x_{t})\|^2 \\
			=&(1+ \frac{1}{2K-1} + 6KL^2\eta^2)\mathbb{E}\|x_{t,k-1}^i-x_{t}\|^2 \\
			&\ + \eta^2\sigma_L^2+ 6K\eta^2\sigma_G^2 + 6K\eta^2\| \nabla f(x_{t})\|^2 \\
			\overset{(a_4)}{\leq}& (1+ \frac{1}{K-1})\mathbb{E}\|x_{t,k-1}^i-x_{t}\|^2 + \eta^2\sigma_L^2 \\
			&\ + 6K\eta^2\sigma_G^2 + 6K\eta^2\| \nabla f(x_{t})\|^2 \\
			\leq& (K-1)[(1+\frac{1}{K-1})^k-1](\eta^2\sigma_L^2+ 6K\eta^2\sigma_G^2 \\
			& \ + 6K\eta^2\| \nabla f(x_{t})\|^2),
	\end{align*}
	where ($a_1$) follows Eq.(\ref{eq:meanequal}), ($a_2$) follows $\|a+b\|^2 \leq (1+\gamma)\|a\|^2+(1+\gamma^{-1})\|b\|^2$ and $\|\sum_{i=1}^n a_i\|^2 \leq n\sum_{i=1}^n \|a_i\|^2$, ($a_3$) follows Eq.(\ref{eq:localvar}), Eq.(\ref{eq:smooth}) and Eq.(\ref{eq:globalvar}), ($a_4$) follows $\eta \leq \frac{1}{4LK}$.
	Thus,
	\begin{equation}\label{eq:multiterm1}
		\begin{aligned}
			&\sum_{k=0}^{K-1}\mathbb{E}\| x_{t,k}^i - x_t\|^2\\
			\leq& \sum_{k=0}^{K-1} \Big( (K-1)[(1+\frac{1}{K-1})^k-1](K-1) \\
			&\ \cdot(\eta^2\sigma_L^2+ 6K\eta^2\sigma_G^2 + 6K\eta^2\| \nabla f(x_{t})\|^2)\Big) \\
			\overset{(a_5)}{\leq}& 5K^2(\eta^2\sigma_L^2+ 6K\eta^2\sigma_G^2 + 6K\eta^2\| \nabla f(x_{t})\|^2) \\
			=& H_1 + H_2\| \nabla f(x_{t})\|^2
		\end{aligned}
	\end{equation}
	where ($a_5$) follows $(1+\frac{1}{K-1})^K \leq 5$, 
	\begin{equation}
		\begin{aligned}
			H_1 =& 5K^2\eta^2\sigma_L^2+ 30K^3\eta^2\sigma_G^2 \\
			H_2 =& 30K^3\eta^2.
		\end{aligned}
	\end{equation}
\end{IEEEproof}

\begin{figure*}[ht]
	\centering
	\subfloat[round-robin]{
		\includegraphics[width=0.8\columnwidth]{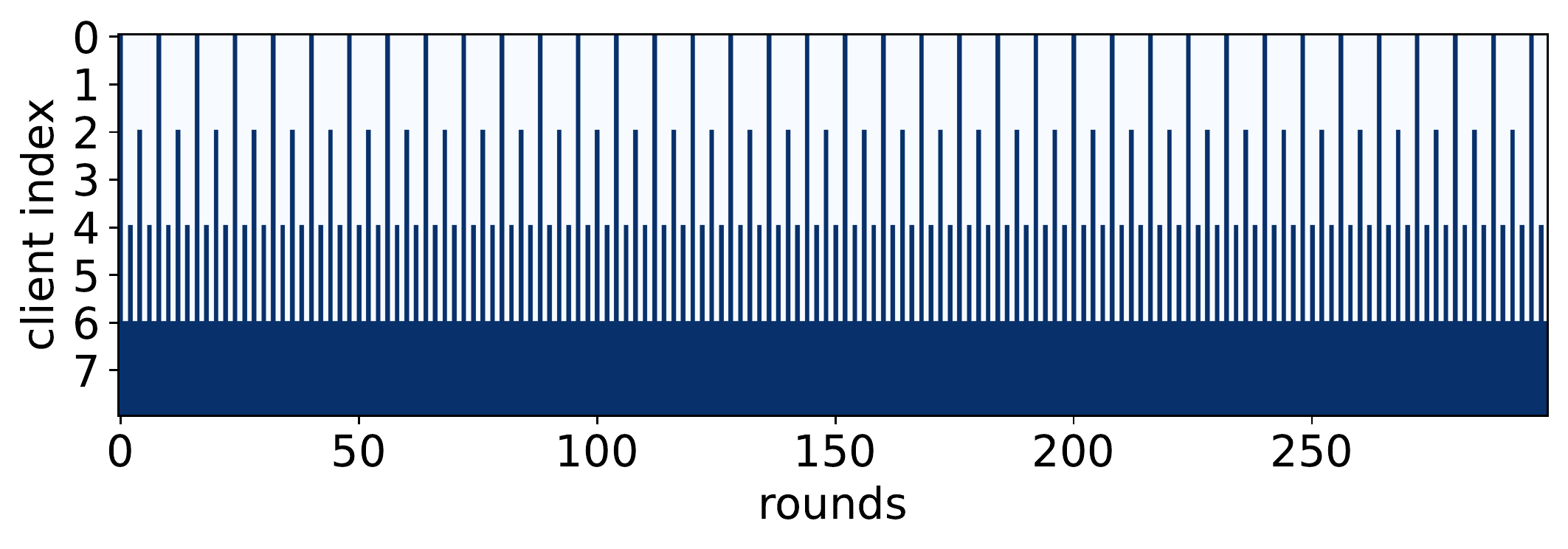} 
		\label{fig:8client_robin} 
	}
	\quad
	\subfloat[ad-hoc]{
		\includegraphics[width=0.8\columnwidth]{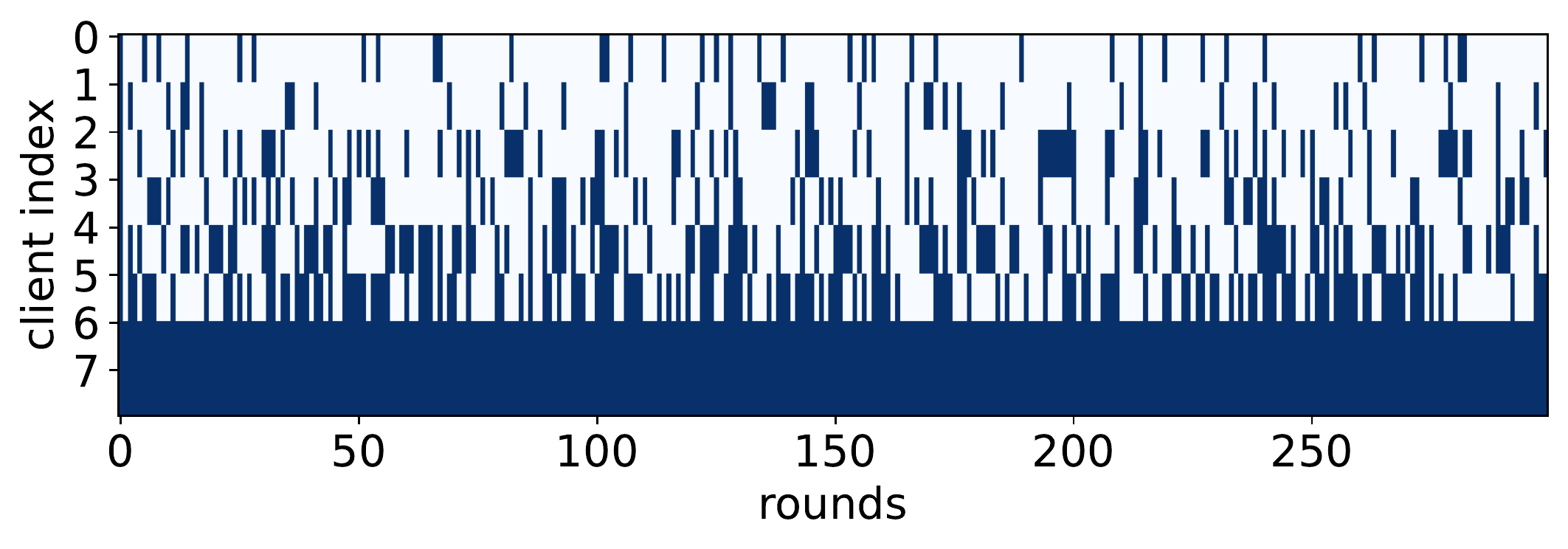} 
		\label{fig:8client_adhoc} 
		
	}
	\caption{The visualization of participation during training process with cross-silo setting.}
	\label{fig:8worker} 
\end{figure*}

\begin{figure*}[ht]
	\centering
	\subfloat[FedAvg with participant ratio 10\%]{
		\includegraphics[width=0.8\columnwidth]{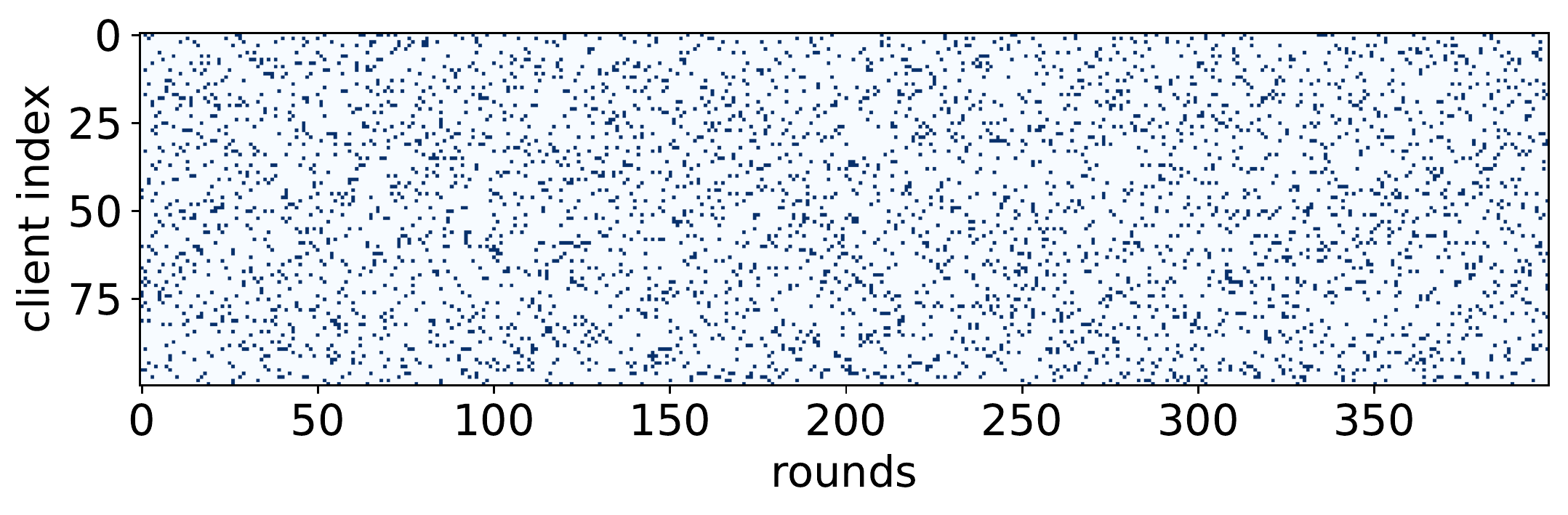} 
		\label{fig:100client01ratio1} 
	}
	\quad
	\subfloat[CC-FedAvg with participant ratio 10\%]{
		\includegraphics[width=0.8\columnwidth]{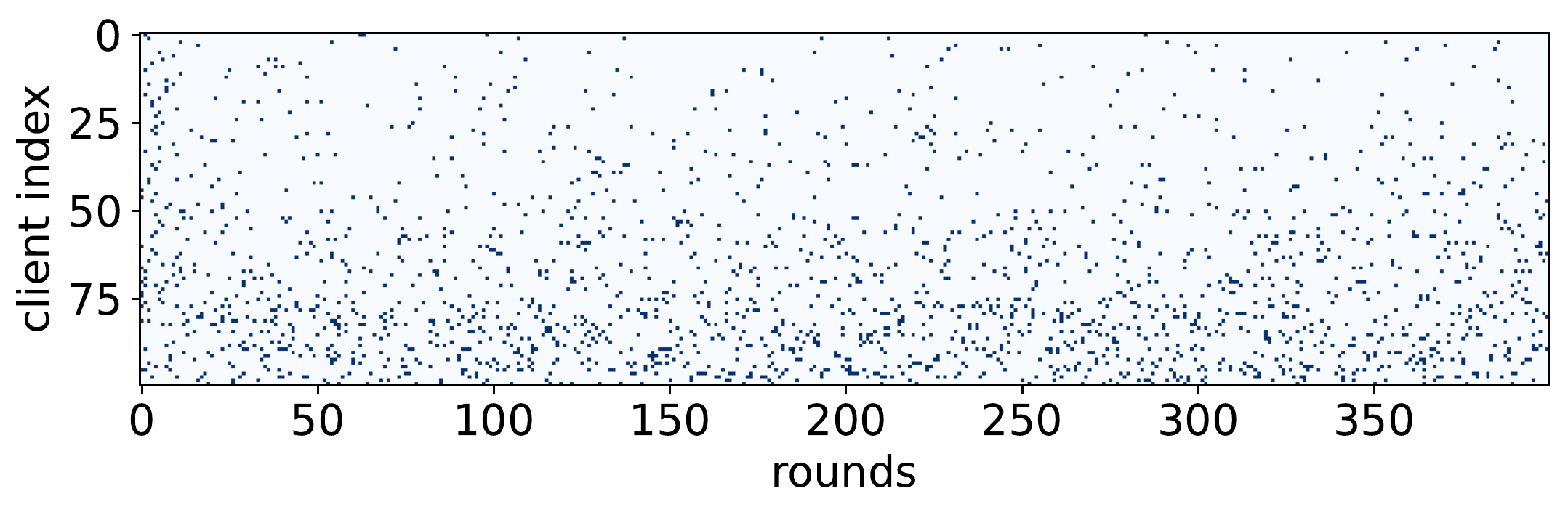} 
		\label{fig:100client01ratio2} 
	}
	\quad
	\subfloat[FedAvg with participant ratio 20\%]{
		\includegraphics[width=0.8\columnwidth]{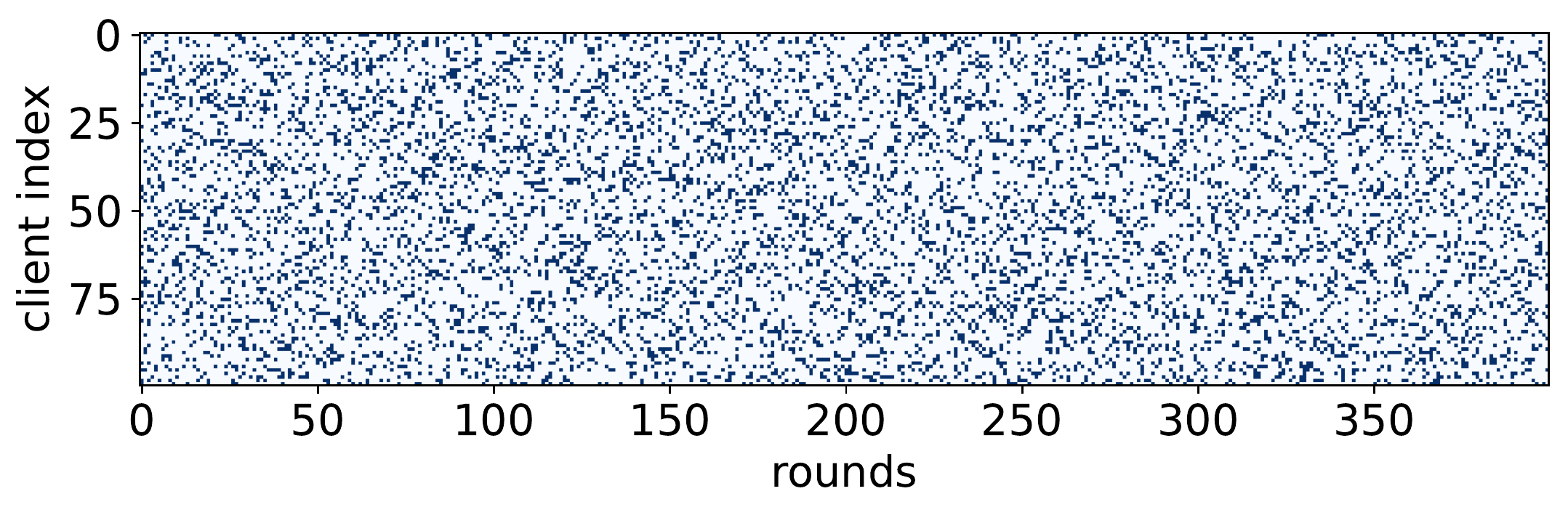} 
		\label{fig:100client02ratio1} 
	}
	\quad
	\subfloat[CC-FedAvg with participant ratio 20\%]{
		\includegraphics[width=0.8\columnwidth]{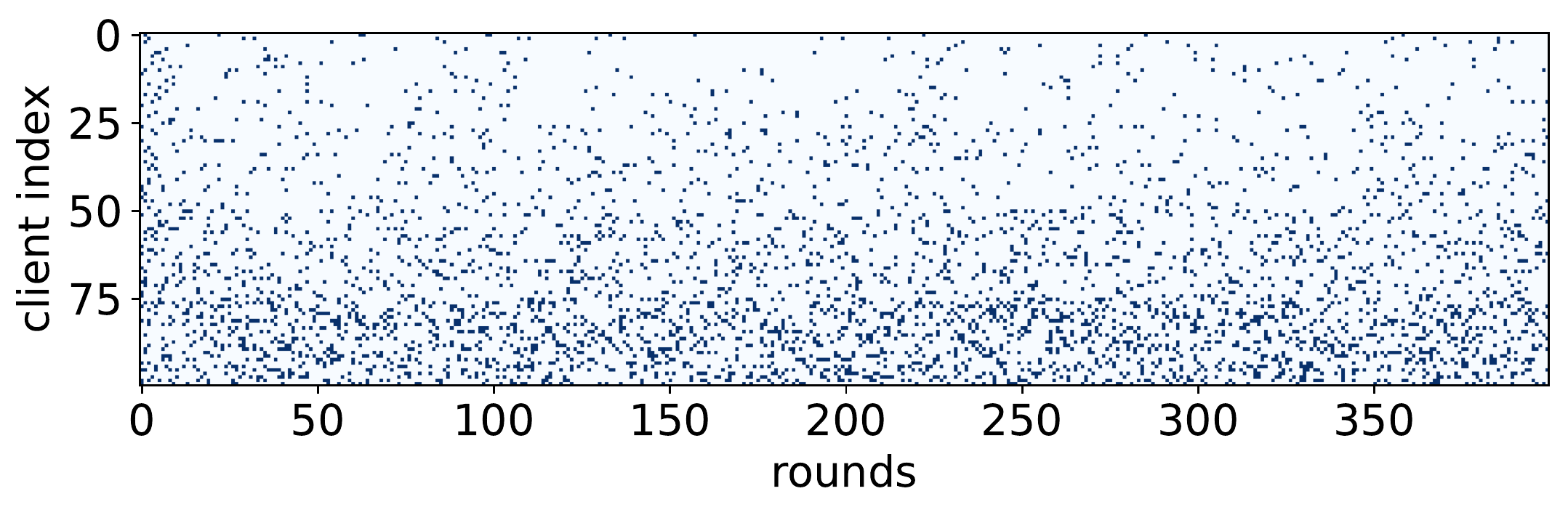} 
		\label{fig:100client02ratio2} 
	}
	\quad
	\subfloat[FedAvg with participation ratio 30\%]{
		\includegraphics[width=0.8\columnwidth]{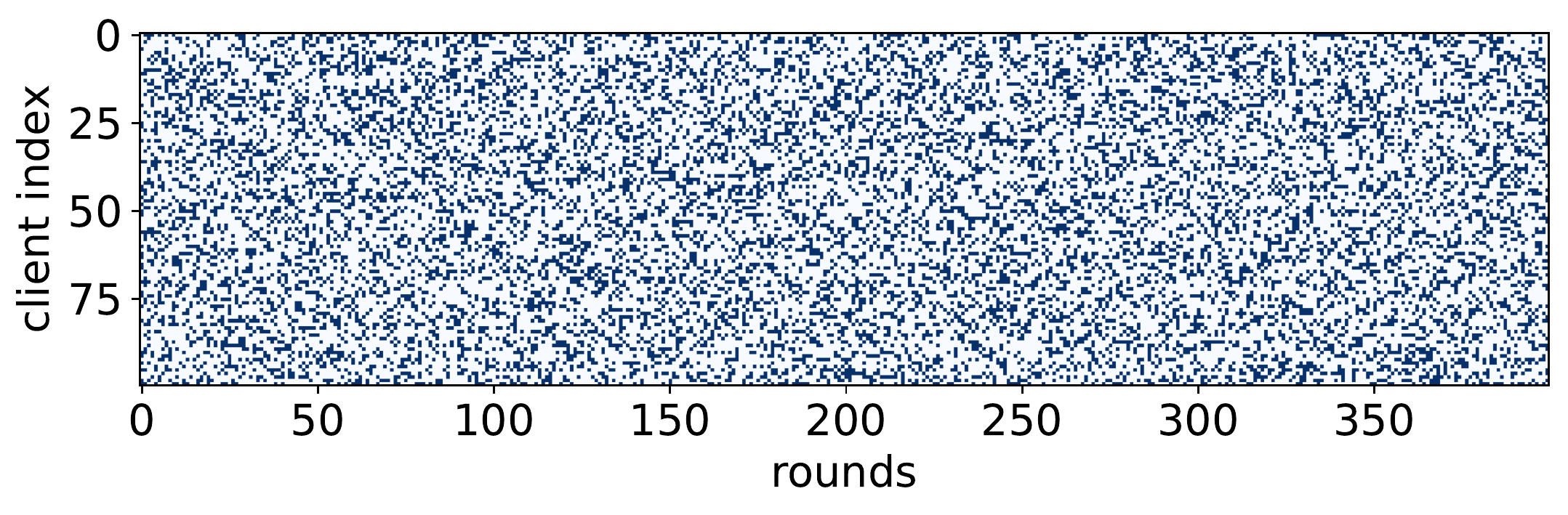} 
		\label{fig:100client03ratio1} 
	}
	\quad
	\subfloat[CC-FedAvg with participant ratio 30\%]{
		\includegraphics[width=0.8\columnwidth]{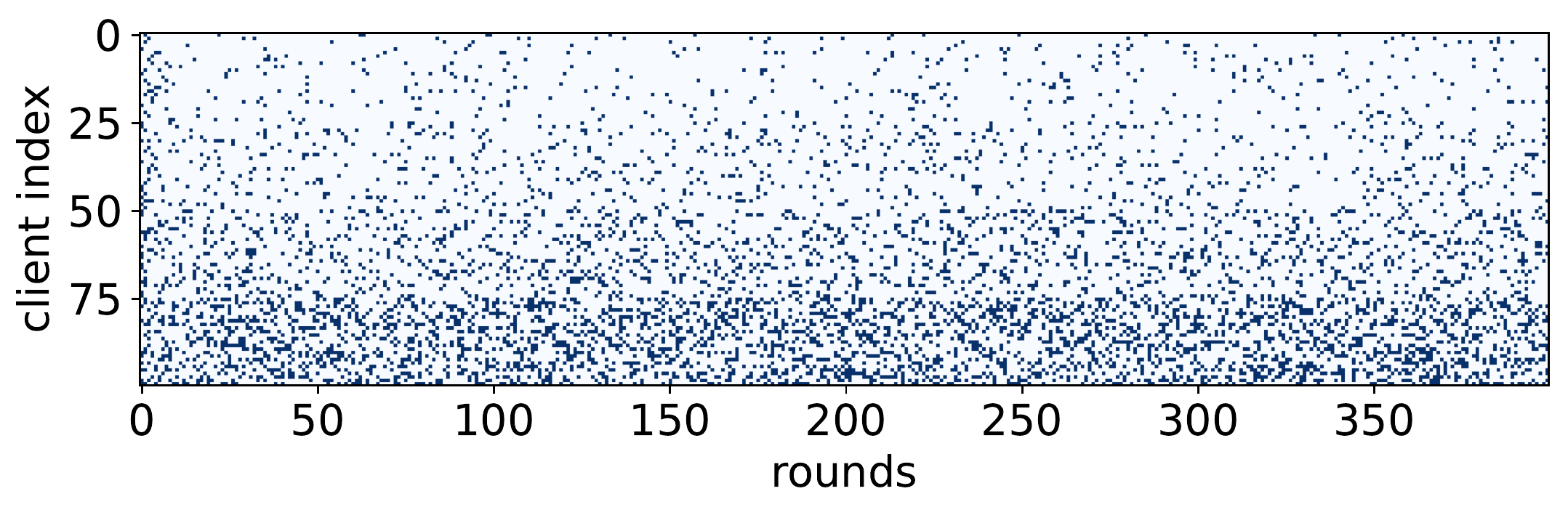} 
		\label{fig:100client03ratio2} 
	}
	\quad
	\subfloat[FedAvg with participant ratio 40\%]{
		\includegraphics[width=0.8\columnwidth]{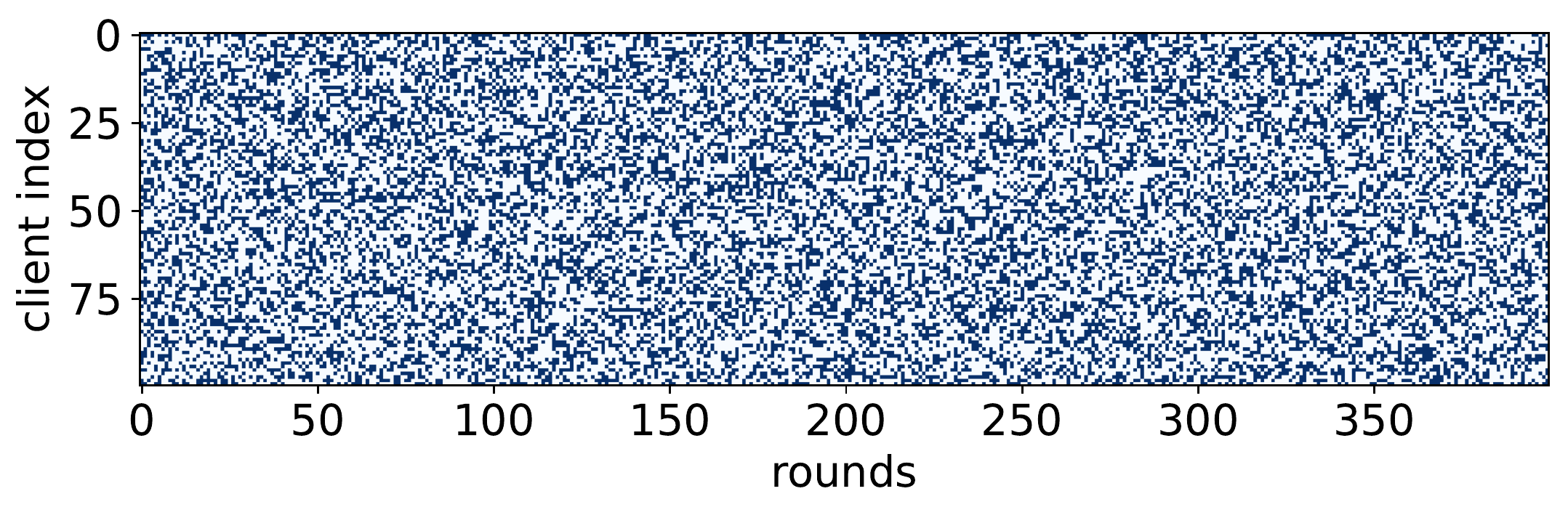} 
		\label{fig:100client04ratio1} 
	}
	\quad
	\subfloat[CC-FedAvg with participant ratio 40\%]{
		\includegraphics[width=0.8\columnwidth]{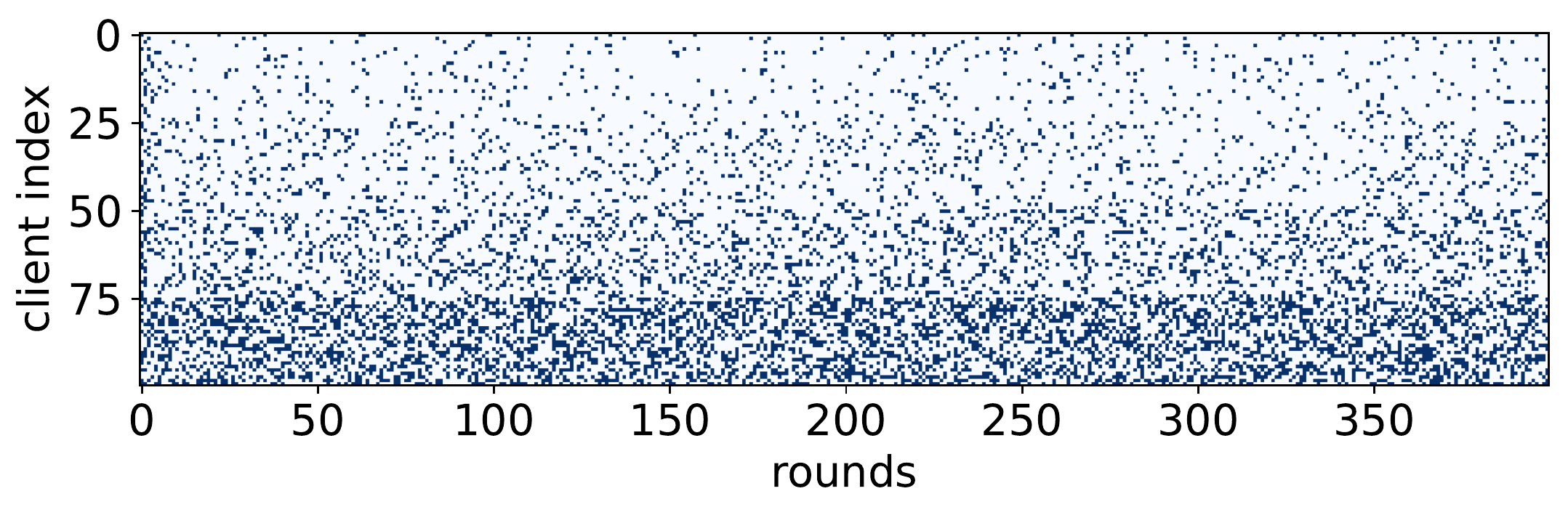} 
		\label{fig:100client04ratio2} 
	}
	\caption{The visualization of participation during training process with cross-device setting.}
	\label{fig:100client} 
\end{figure*}

For convenient we denote $M=r\cdot N$. Suppose in round $t$, there are up to $M$ clients skipping the iterations (since $M$ can be selected from the whole rounds, we ignore the subscript $t$ and regard it as a constant for simplicity), i.e. they perform line~15 
in Algorithm 1. Moreover, the clients that continuously skip iterations exist. Without loss of generality, we suppose $\mathcal{M}_j (0 \leq j\leq W)$ denotes the client set in which the clients have continuously skipped $j$ rounds of local iteration. For instance, for the client $i$ in $\mathcal{M}_2$, it means $\Delta_{t}^i=\Delta_{t-1}^i=\Delta_{t-2}^i$. Particularly, the clients in $\mathcal{M}_0$ performs local SGD iterations in this round. Obviously, $\forall i\neq j, \mathcal{M}_i \cap \mathcal{M}_j = \emptyset$. 
Suppose $m_j = \mathbb{E}_t[|\mathcal{M}_j|] (0 \leq j\leq W)$ (for the clients, if the probability of skipping training are the same for each round, $m_j$ is a constant in expectation), it has $\sum_{j=0}^{W} m_j = N$ and  $ \sum_{j=1}^{W} m_j = M$. Based on this condition, we have
\begin{equation}
	\begin{aligned}
		\Delta_t &= \frac{1}{N}\sum_{i}^N \Delta_t^i = \frac{1}{N}\sum_{j=0}^W\sum_{i \in \mathcal{M}_j} \Delta_{t-j}^i \\&= -\frac{1}{N}\sum_{j=0}^W\sum_{i \in \mathcal{M}_j}\sum_{k=0}^{K-1} \eta g_{t-j,k}^i.
	\end{aligned}
\end{equation}

\begin{lemma}\label{lemma:delta}
	\textit{For the moving vector of global model $\Delta_t$, we have 
	\begin{equation}
		\begin{aligned}
			\mathbb{E}\|\Delta_t\|^2 \leq \|\mathbb{E}[\Delta_t]\|^2 + \frac{K\eta^2}{N}\sigma_L^2.
		\end{aligned}
	\end{equation}}

	
\end{lemma}

\begin{IEEEproof}
	\begin{equation}
		\begin{aligned}
			\mathbb{E}[\Delta_t] &= -\frac{1}{N}\sum_{j=0}^W\sum_{i \in \mathcal{M}_j}\sum_{k=0}^{K-1} \eta \nabla f_i(x_{t-j,k}^i), \\
		\end{aligned}
	\end{equation}
	Thus 
	\begin{equation}
		\begin{aligned}
			\mathbb{E}\|\Delta_t\|^2 &=  \mathbb{E}\|\Delta_t - \mathbb{E}[\Delta_t] + \mathbb{E}[\Delta_t]\|^2 \\&= \mathbb{E}\|\mathbb{E}[\Delta_t]\|^2 + \mathbb{E}\|\Delta_t - \mathbb{E}[\Delta_t]\|^2 \\
			&= \|\mathbb{E}[\Delta_t]\|^2 \\
			&\ + \|\frac{1}{N}\sum_{j=0}^W\sum_{i \in \mathcal{M}_j}\sum_{k=0}^{K-1} \eta (\nabla f_i(x_{t-j,k}^i) - g_{t-j,k}^i) \|^2\\
			&\leq \|\mathbb{E}[\Delta_t]\|^2 + \frac{K\eta^2}{N}\sigma_L^2.
		\end{aligned}
	\end{equation}
\end{IEEEproof}

\section{Visualization of Participation}\label{visualization}

Figure~\ref{fig:8worker} visualizes the participation of clients during training process of Table~\ref{table-silo-cifar} with round-robin and ad-hoc schedules, respectively. Client $y$ participates (does not participate) in round $x$ is shown by the dark (light) color at $(x,y)$. From the figure, we can intuitively see the difference of participation rounds across clients, and CC-FedAvg spends much fewer computation resources than vanillar federated learning (in this situation all the points in the figure are dark).

Figure~\ref{fig:100client} visualizes the participation of clients during training process of Table~\ref{table-device-fmnist1}, Table~\ref{table-device-fmnist2} and Table~\ref{table-device-fmnist3}. Due to cross-device setting, the participation of one client in CC-FedAvg is determined by both sever selection and client decision (on the contrary, the participation of one client in FedAvg is only determined by sever selection).  Fig.~\ref{fig:100client01ratio1} and Fig.~\ref{fig:100client01ratio2} visualize the actual participation in training where the server selects 10\% clients in each round. It is obvious that CC-FedAvg needs much fewer participation than FedAvg. The similar situations are shown from Fig.~\ref{fig:100client02ratio1} to Fig.~\ref{fig:100client04ratio2}.

\section{Extra Experiments under Cross-device Settings}


\begin{table*}[htbp]
	\centering
	\caption{Performance (top-1 test accuracy) comparison on FMNIST with different data heterogeneity, where $N=100$, $\beta=4$, and the classes of training data are highly skewed across clients with different computation resources.} 
	\label{table-device-fmnist2}
	\vskip 0.15in
	\begin{center}
		\begin{small}
			\begin{sc}
	\begin{tabular}{|c|c|c|c|c|c|}
			\hline
			& 10\% & 20\% & 30\% & 40\%  \\	\hline
			FedAvg     & 67.78$\pm$7.35  & 70.51$\pm$ 4.68 & 75.25$\pm$ 2.51 & 76.86$\pm$ 2.26 
			\\\hdashline[1pt/1pt]
			Strategy 1 & 55.30$\pm$ 9.11 & 58.59$\pm$ 8.56 & 58.60$\pm$6.54 & 60.17$\pm$5.68  \\\hline
			Strategy 2 & 52.40$\pm$8.29 & 60.24$\pm$ 6.69 & 69.37$\pm$ 3.37 & 71.75$\pm$3.14   \\\hline
			CC-FedAvg & \textbf{58.47$\pm$9.86} & \textbf{66.11$\pm$6.32} &\textbf{70.99$\pm$ 4.19} & \textbf{73.89$\pm$3.62} \\ 
			\hline
		\end{tabular}
	\end{sc}
\end{small}
\end{center}
\vskip -0.1in
\end{table*}

\begin{table*}[htbp]
	\centering
	\caption{Performance (top-1 test accuracy) comparison on FMNIST with different data heterogeneity, where $N=100$, $\beta=4$, and the classes of training data are moderately skewed across clients with different computation resources.}
	\label{table-device-fmnist3} 
		\vskip 0.15in
	\begin{center}
		\begin{small}
			\begin{sc}
	\begin{tabular}{|c|c|c|c|c|c|}
			\hline
			& 10\% & 20\% & 30\% & 40\%  \\	\hline
			FedAvg     & 74.52$\pm$3.63  & 80.14$\pm$ 2.07 & 81.03$\pm$ 1.86 & 81.62$\pm$ 1.11 
			\\\hdashline[1pt/1pt]
			Strategy 1 & 65.64$\pm$ 5.77 & 74.92$\pm$ 4.40 & 77.53$\pm$2.89 & 78.29$\pm$2.37  \\\hline
			Strategy 2 & 59.16$\pm$5.88 & 71.26$\pm$ 4.03 & 75.17$\pm$ 3.12 & 76.97$\pm$1.67   \\\hline
			CC-FedAvg & \textbf{70.14$\pm$4.05} & \textbf{78.40$\pm$2.17} &\textbf{79.30$\pm$ 2.90} & \textbf{80.36$\pm$1.61} \\ 
			\hline
		\end{tabular}
	\end{sc}
\end{small}
\end{center}
\vskip -0.1in	
\end{table*}

\begin{figure}[ht]
	\centering
	\subfloat[Strategy 1]{
		\includegraphics[width=0.8\columnwidth]{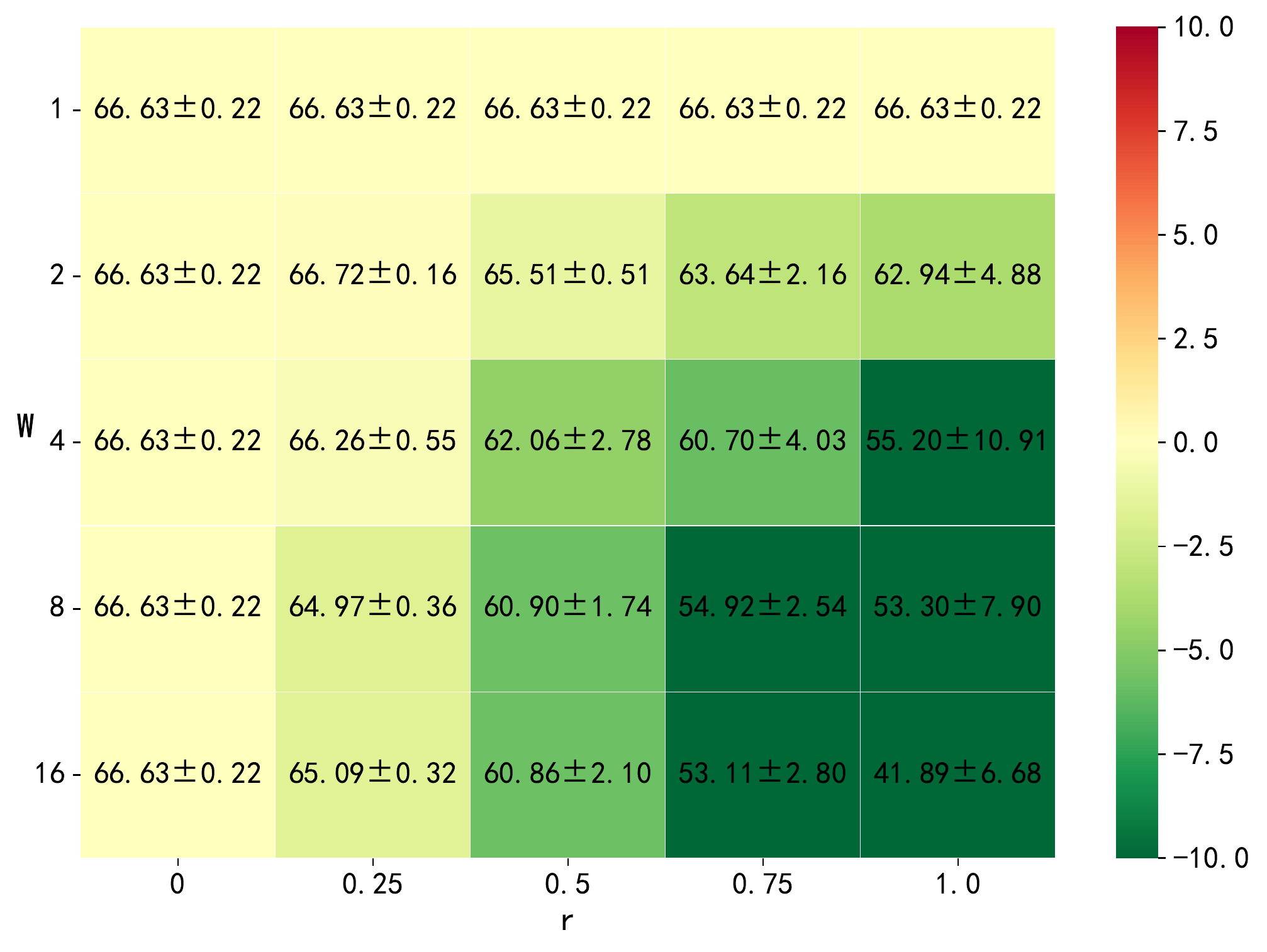} 
		\label{fig:heatmap_cifar10_silo_md7} 
	}
	\quad
	\subfloat[Strategy 2]{
		\includegraphics[width=0.8\columnwidth]{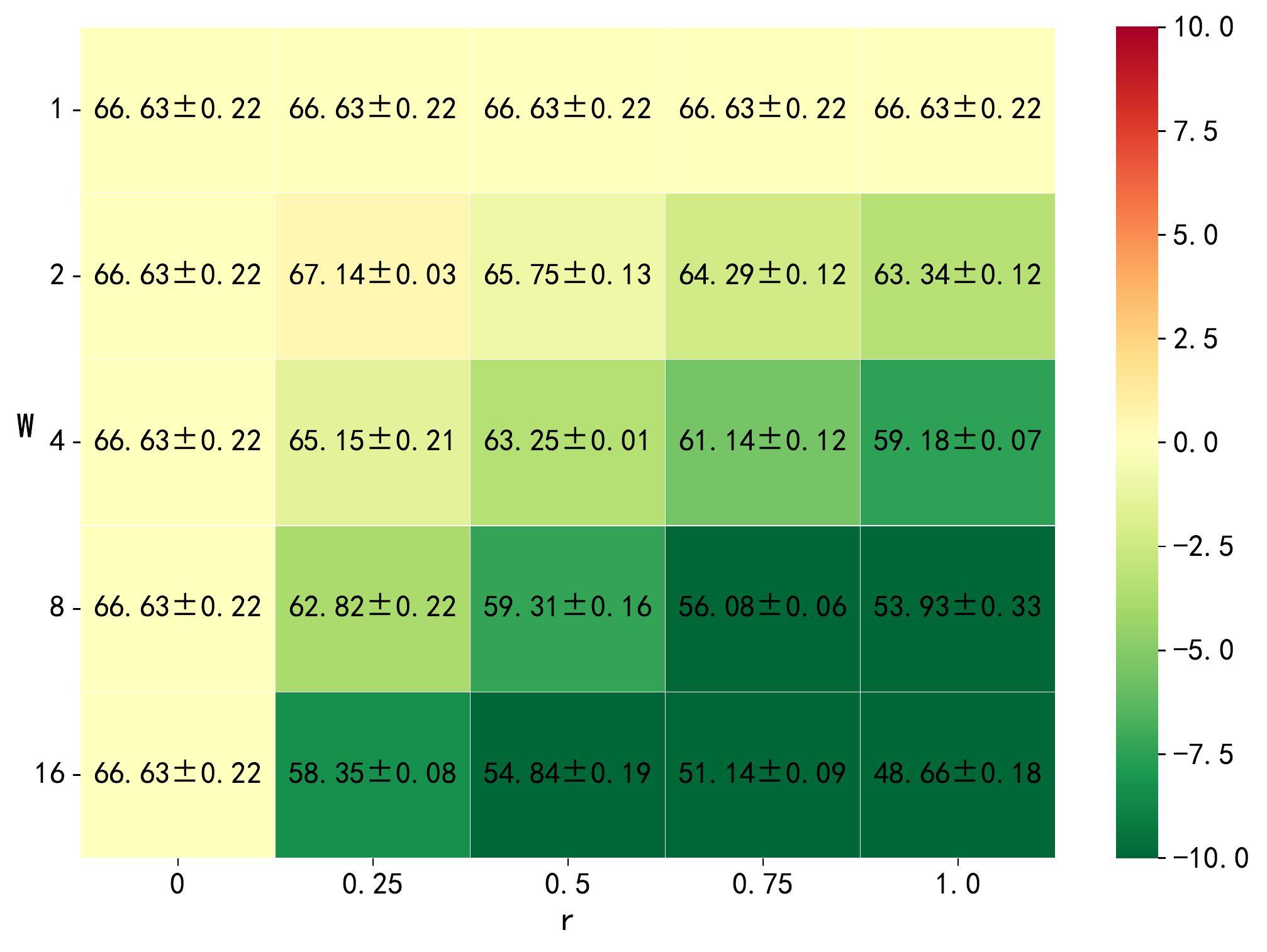} 
		\label{fig:heatmap_cifar10_silo_md5} 
	}
	\caption{The performance changes with varying of $r$ and $W$ under cross-silo scenarios (CIFAR-10).}
	\label{fig:MandW_cifar10} 
\end{figure}

\begin{figure}[htbp]
	\centering
	\subfloat[Strategy 1]{
		\includegraphics[width=0.8\columnwidth]{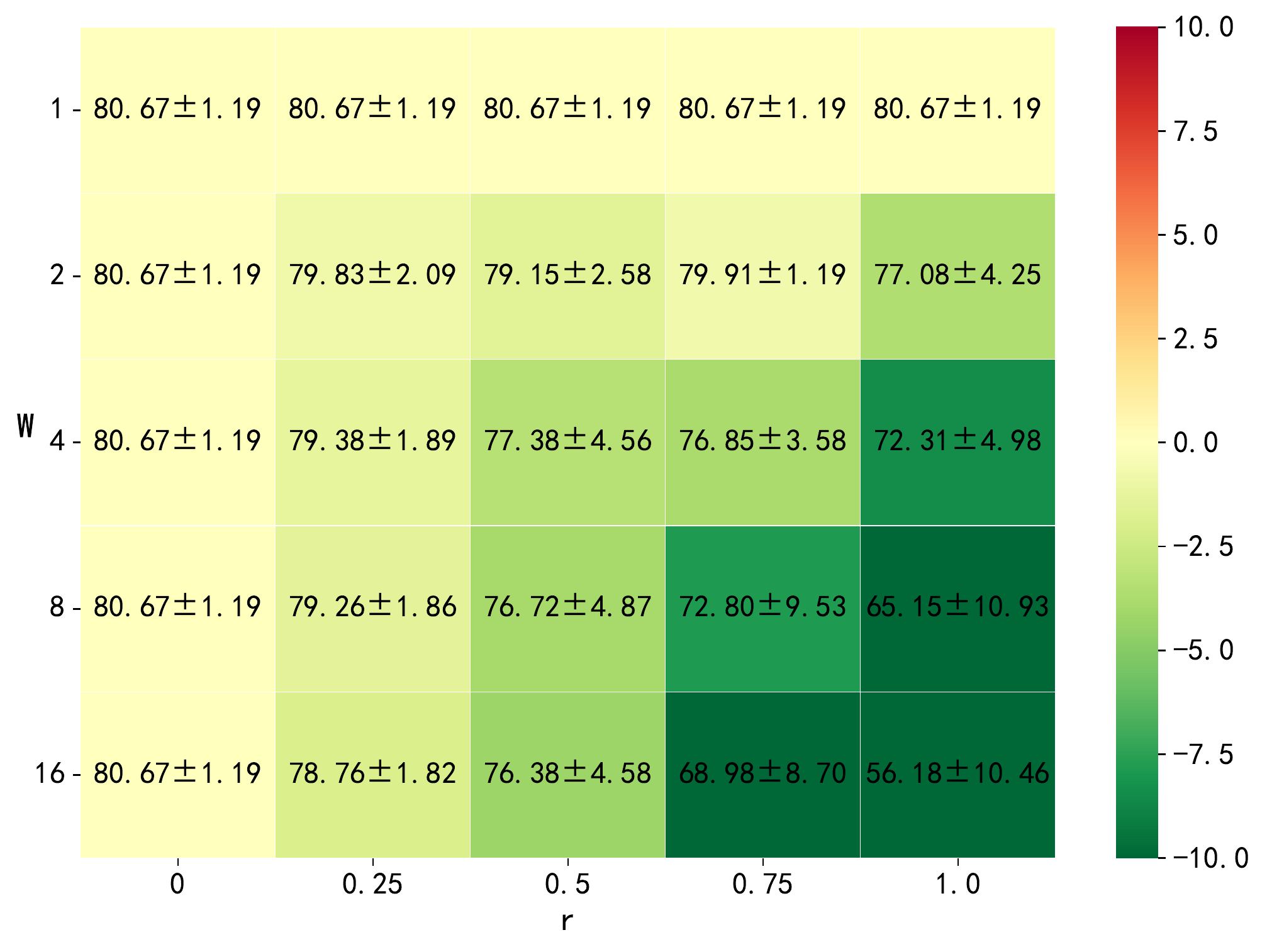} 
		\label{fig:heatmap_fmnist_device_md7} 
	}
	\quad
	\subfloat[Strategy 2]{
		\includegraphics[width=0.8\columnwidth]{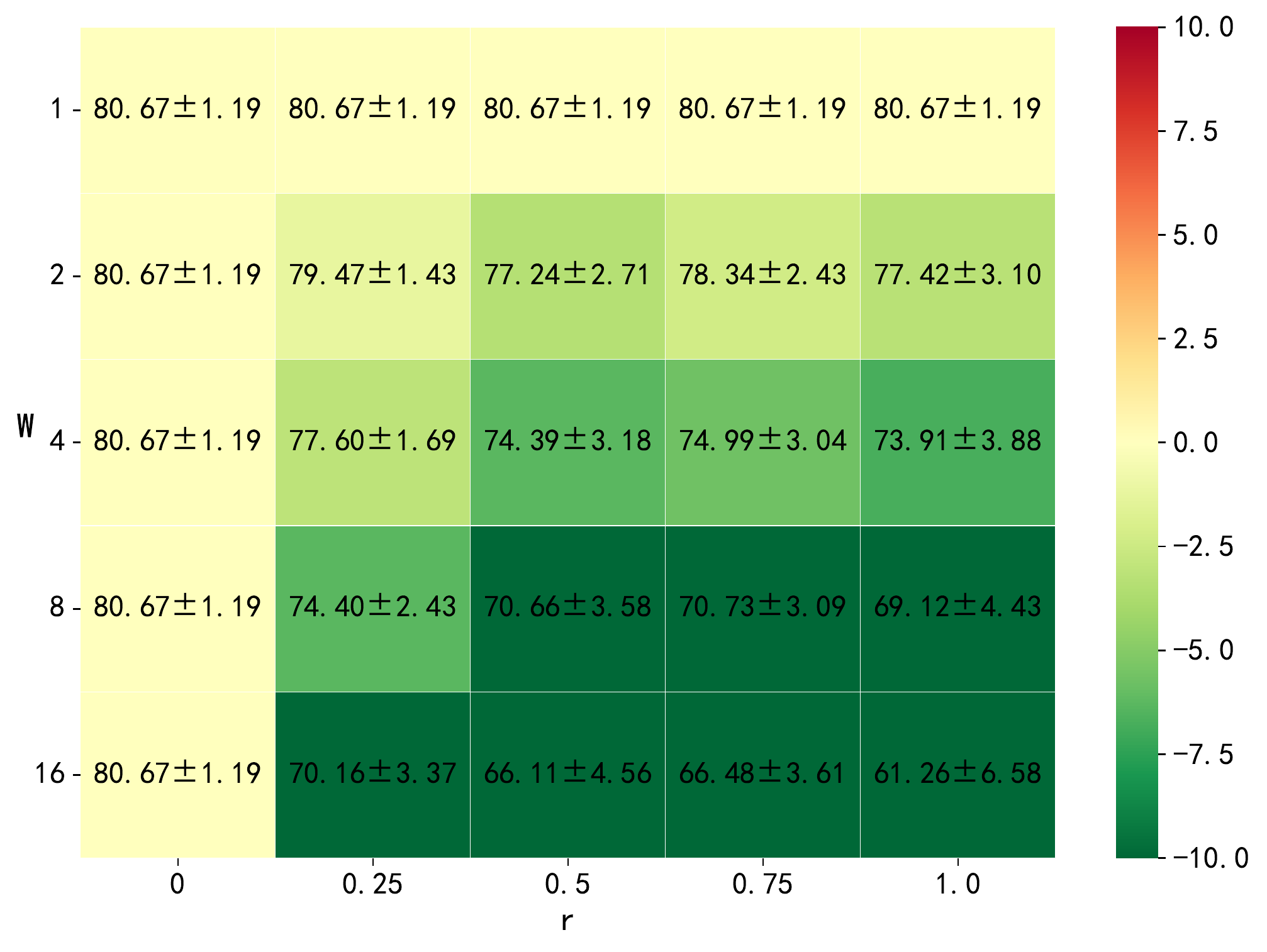} 
		\label{fig:heatmap_fmnist_device_md5} 
	}
	\caption{The performance changes with varying of $r$ and $W$ under cross-device scenarios (FMNIST).}
	\label{fig:MandW_fmnist} 
\end{figure}
In Table~\ref{table-device-fmnist1} each client is randomly assigned two types of data. Each class of data is spread evenly among 10 clients. We assign $p_i$ at random (i.e., we randomly assign the computation resource budgets). Each type of data is present on both the clients with adequate resources and the clients with insufficient resources. Therefore, the data distribution is not skewed during training. 
In practice, missing some clients may skew the real-time training data. Table~\ref{table-device-fmnist2} and Table~\ref{table-device-fmnist3} study the impact, respectively. In Table~\ref{table-device-fmnist2} we assign two classes of data two 20 clients randomly, and assign the same $p_i$ to the clients with the same class of data. As a consequence, each class of data is either on the client with adequate resources or the clients with insufficient resources. The data distribution is significantly skewed during training. 
Compared with Table~\ref{table-device-fmnist1}, in this situation the performance of CC-FedAvg deteriorates more. But compared with the other two strategies, this method is much more robust.
For example, when participant ratio is  20\%, compared to FedAvg, CC-FedAvg reduces the performance by only 4\%, while the performance is deteriorated by more than 10\% by another two strategies. Furthermore, in Table~\ref{table-device-fmnist1} Strategy 1 is better than Strategy 2 in most cases, but the result is reversed in Table~\ref{table-device-fmnist1}. On the contrary, CC-FedAvg is consistent under the varying of settings. 
Table~\ref{table-device-fmnist3} also verifies the advantages of CC-FedAvg, which is a compromise between Table~\ref{table-device-fmnist1} and Table~\ref{table-device-fmnist2}, with only 10\% of clients using the setting of Table~\ref{table-device-fmnist2} (these clients have adequate resources) and the rest following the setting of Table~\ref{table-device-fmnist1}. Correspondingly, the data distribution is moderately skewed during training.

\section{Influence of $r$ and $W$}\label{sub:influence}

To compare with baselines, we also investigate the model performance of Strategy~\ref{strategy1} and Strategy~\ref{strategy2} with different $r$ and $W$. Fig.~\ref{fig:MandW_cifar10} corresponds to Fig.~\ref{fig:heatmap_cifar10_silo_md4}, and Fig.~\ref{fig:MandW_fmnist} corresponds to Fig.~\ref{fig:heatmap_fmnist_device_md4}. We can find CC-FedAvg performs stable under most $r$ and $W$, while Strategy~\ref{strategy1} and Strategy~\ref{strategy2} degrade severely with moderately large $r$ or $W$.

\bibliographystyle{plain}
\bibliography{ref}

\end{document}